\pdfoutput=1

\documentclass[11pt]{article}

\usepackage[]{ACL2023}

\usepackage{times}
\usepackage{latexsym}

\usepackage[T1]{fontenc}

\usepackage[utf8]{inputenc}

\usepackage{microtype}

\usepackage{inconsolata}
\usepackage{xcolor,pifont}
\newcommand*\colourcheck[1]{%
  \expandafter\newcommand\csname #1check\endcsname{\textcolor{#1}{\ding{52}}}%
}

\makeatletter
\renewcommand\@makefntext[1]{%
    \parindent 0em%
    \noindent
    \hb@xt@1.8em{\hss\@makefnmark}#1}
\makeatother
\usepackage[hang]{footmisc}

\colourcheck{blue}
\colourcheck{green}
\colourcheck{red}

\usepackage[utf8]{inputenc} 
\usepackage[T1]{fontenc}    
\usepackage{hyperref}       
\usepackage{url}            
\usepackage{booktabs}       
\usepackage{amsfonts}       
\usepackage{nicefrac}       
\usepackage{microtype}      
\usepackage{xcolor}         
\usepackage{listings}
\usepackage{algorithm}
\usepackage{algorithmic}
\usepackage{graphicx}
\usepackage{multicol}
\usepackage{CJKutf8}
\usepackage{booktabs}
\usepackage{subcaption}
\usepackage[utf8]{inputenc}
\usepackage{titlesec}
\usepackage[normalem]{ulem}
\usepackage{xspace}
\usepackage{mathtools}
\usepackage{multirow}
\usepackage{graphicx}
\usepackage{subcaption}
\usepackage{pythonhighlight}
\usepackage{wrapfig,lipsum,booktabs}
\usepackage{listings}
\usepackage{listings}
\definecolor{codegreen}{rgb}{0,0.6,0}
\definecolor{codegray}{rgb}{0.5,0.5,0.5}
\definecolor{codepurple}{rgb}{0.58,0,0.82}
\definecolor{backcolour}{rgb}{0.95,0.95,0.92}

\lstdefinestyle{mypython}{
    backgroundcolor=\color{backcolour},   
    commentstyle=\color{codegreen},
    keywordstyle=\color{magenta},
    numberstyle=\tiny\color{codegray},
    stringstyle=\color{codepurple},
    basicstyle=\ttfamily\footnotesize,
    breakatwhitespace=false,         
    breaklines=true,                 
    captionpos=b,                    
    keepspaces=true,                 
    numbers=left,                    
    numbersep=5pt,                  
    showspaces=false,                
    showstringspaces=false,
    showtabs=false,                  
    tabsize=2
}

\usepackage[font=small]{caption}
\usepackage{amsthm}
\usepackage{tabularray}
\usepackage[nameinlink]{cleveref}
\crefformat{section}{\S#2#1#3} 
\crefformat{appendix}{\S#2#1#3} 
\crefname{algorithm}{Alg.}{Algs.}
\crefformat{subsection}{\S#2#1#3}
\Crefname{equation}{Eq.}{Eqs.}
\Crefname{figure}{Fig.}{Figs.}
\usepackage[colorinlistoftodos,prependcaption,textsize=tiny]{todonotes}

\usepackage{caption}
\usepackage{soul}

\usepackage{float}
\usepackage{array}

\usepackage{multirow}
\usepackage{hhline}
\usepackage{microtype}
\usepackage{tcolorbox}

\usepackage{inconsolata}

\usepackage{microtype}
\usepackage{hyperref}
\usepackage{booktabs}
\usepackage{graphicx}
\usepackage{footmisc}
\usepackage{subcaption,siunitx,booktabs}
\usepackage{amsthm}
\usepackage{amsmath}
\newtheorem{theorem}{Theorem}[section]

\newtheorem{definition}{Definition}[section]

\usepackage[nameinlink]{cleveref}
\crefformat{section}{\S#2#1#3} 
\crefname{algorithm}{Alg.}{Algs.}
\crefname{table}{Tab.}{Tab.}
\crefname{theorem}{Theorem}{Theorem}
\crefformat{theorem}{Thm. #2#1#3}
\crefformat{assum}{Asm. #2#1#3}
\crefformat{appendix}{Appx. #2#1#3}
\crefformat{hypothesis}{Hyp. #2#1#3}
\crefformat{subsection}{\S#2#1#3}
\Crefname{equation}{Eq.}{Eqs.}
\Crefname{figure}{Fig.}{Figs.}

\definecolor{inc}{RGB}{84,123,71}
\definecolor{dec}{RGB}{219, 48, 122}
\usepackage{inconsolata}
\usepackage{url}            
\usepackage{booktabs}       
\usepackage{amsfonts}       
\usepackage{nicefrac}       
\usepackage{microtype}      
\usepackage{xcolor}         

\usepackage{xcolor, soul,todonotes} 
\definecolor{kmycolor}{rgb}{0.858, 0.188, 0.878}

\title{\emph{LLMs Are Biased Towards Output Formats!}\\Systematically Evaluating and Mitigating Output Format Bias of LLMs}

\author{Do Xuan Long$^{1,2}$, Hai Nguyen Ngoc$^{3}$, Tiviatis Sim$^{1,4}$, Hieu Dao$^{1}$, Shafiq Joty$^{5,6}$,\\
\textbf{Kenji Kawaguchi$^{1}$, Nancy F. Chen$^{2}$, Min-Yen Kan$^{1}$} \\
$^{1}$National University of Singapore, $^{2}$Institute for Infocomm Research (I$^2$R), A*STAR,\\ $^{3}$VinAI Research, $^{4}$Institute of High Performance Computing (IHPC), A*STAR,\\ $^{5}$Salesforce Research, $^{6}$Nanyang Technological University\\
\small{\{xuanlong.do, tiviatis\}@u.nus.edu}, \small{haibeo2552001@gmail.com
},\\
\small{sjoty@salesforce.com}, \small{\{daohieu, kenji, kanmy\}@comp.nus.edu.sg}, \small{nfychen@i2r.a-star.edu.sg}
}

\begin{document}
\maketitle
\begin{abstract}
We present the first systematic evaluation examining format bias in \textbf{performance} of large language models (LLMs).
Our approach distinguishes between two categories of an evaluation metric under format constraints to reliably and accurately assess performance: one measures performance when format constraints are adhered to, while the other evaluates performance regardless of constraint adherence. We then define a metric for measuring the format bias of LLMs and establish effective strategies to reduce it. Subsequently, we present our empirical format bias evaluation spanning four commonly used categories---multiple-choice question-answer, wrapping, list, and mapping---covering $15$ widely-used formats. Our evaluation on eight generation tasks uncovers significant format bias across state-of-the-art LLMs. We further discover that improving the format-instruction following capabilities of LLMs across formats potentially reduces format bias. Based on our evaluation findings, we study prompting and fine-tuning with synthesized format data techniques to mitigate format bias. Our methods successfully reduce the variance in ChatGPT's performance among wrapping formats from $235.33$ to $0.71$ 
($\%^2$). 

\end{abstract}

\section{Introduction}

\begin{figure}[t]
\centering
\includegraphics[width=.9\columnwidth, trim={0cm 0cm 0cm 0cm},clip]{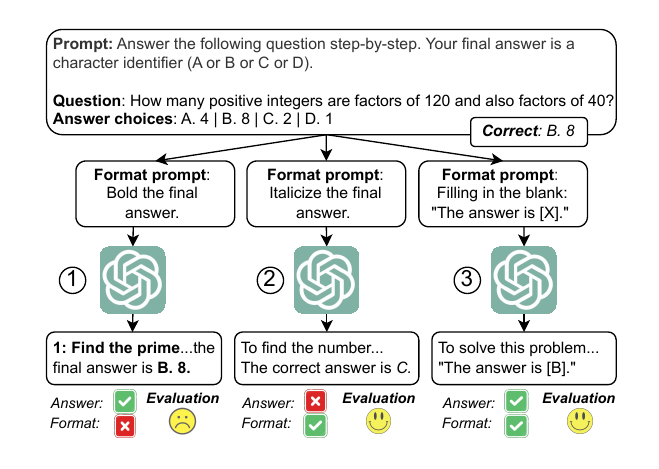}
\caption{\small{A MMLU example \cite{hendrycks2020measuring} with ChatGPT across different formats. {In Case~(1), the model can answer the question but fails to bold only the answer, hindering automatic evaluation. In Case~(2), the model follows the format but produces an incorrect result. In Case~(3), the model yields the correct answer and format.} These show bias in ChatGPT's performance across formats. 
}}
\vspace{-3mm}
\label{fig:formateval-teaser}
\end{figure}

To unlock the full potential of automating real-world applications, state-of-the-art large language models (LLMs) \citep{brown2020language,chowdhery2023palm,openai2022chatgpt,touvron2023llama}  are increasingly leveraged to tailor outputs to specific task formats. This powerful approach has driven advancements across domains including medicine \cite{thirunavukarasu2023large,clusmann2023future}, data analysis \cite{cheng-etal-2023-gpt,liu-etal-2023-jarvix}, and even evaluating models themselves \cite{chiang-lee-2023-large,chang2024survey}. 
Employing LLMs in such applications heavily depends on not only their format-following capability but also \emph{high-quality results within formats}.


While many studies, including those listed above, have utilized LLMs to output in specific formats, understanding their format capabilities is critical yet has received limited attention.  Recently, \citet{zhou2023instructionfollowing} and \citet{xia-etal-2024-fofo} introduced benchmarks assessing LLM format-following proficiency. However, these studies neglect deeper insights into how these formats impact model performance, which is the ultimate concern for industrial and practical applications. Given numerous formats recently introduced across tasks and models, assessing this aspect is essential for business yet challenging. Evaluation can be ambiguous and often overlook cases where models provide correct answers but are formatted wrong (Case~(1) in  \Cref{fig:formateval-teaser}). 


Bridging these gaps, we conduct the first systematic evaluation of the format bias
of LLMs. Our study attempts to answer the research questions:

\vspace{-2mm}
\begin{tcolorbox}[sharp corners, colback=white, boxrule=-1pt]
\textit{How can we systematically and accurately assess format bias in the performance of LLMs, and to what extent are they biased?}
\end{tcolorbox}
\vspace{-2mm}


To fairly assess bias in model performance across formats, it is crucial to evaluate all scenarios depicted in \Cref{fig:formateval-teaser}. Nonetheless, Case~(1) is challenging to automatically measure, requiring costly human investigation. Therefore, we propose a reliable estimator for evaluating LLM performance under format constraints without human intervention by considering format-following scores. We start by redefining LLM evaluation metrics into two distinct classes to construct the estimator, as detailed in \Cref{ssec:format-eval-metrics}.
Accordingly, we define a metric to quantify format bias in LLMs and establish criteria for evaluating methods that successfully mitigate this bias (\Cref{ssec:format-eval-bias}). Based on these formulations, we present our format evaluation framework, comprising of the widely-utilized categories of multiple-choice question--answer (MCQ; \Cref{sec:mcq}), wrapping (\Cref{sec:wrapping}), list (\Cref{sec:list}) and mapping formats (\Cref{sec:mapping}). 

Across $15$ widely-used formats, our evaluation with zero-shot and zero-shot chain-of-thought prompting \cite{kojima2022large} on eight question-answering and reasoning tasks reveals substantial performance and format-instruction following inequalities. To address this, we examine prompting and fine-tuning using synthesized format data techniques that work for both open- and closed-source LLMs. Our study validates that enhancing LLMs' capabilities to follow format instructions potentially mitigates format bias: (1) Prompting with demonstrations and (2) Repeating format instructions substantially alleviates this bias. Moreover, we investigate (3) Synthesizing limited format data based on our evaluation results for fine-tuning. Our approaches significantly decrease ChatGPT performance variance across wrapping formats from $235.33$ to $0.71$ ($\%^2$) on MMLU \cite{hendrycks2020measuring}. Our key contributions are: 

\begin{enumerate}
    \item We introduce the first systematic framework\footnote{Our codes and data will be made publicly available at \href{https://github.com/dxlong2000/FormatEval}{link}.} to evaluate format performance bias in LLMs.
    \item A large-scale evaluation spanning $15$ formats, $8$ tasks, and {$4$} models revealing substantial LLM performance variance across formats.
    \item The development of $3$ novel prompting and fine-tuning methods to mitigate this bias.
\end{enumerate}


\section{Related Works} \label{ssec:related-work}
Large language models (LLMs) have shown remarkable proficiency in formatting outputs to meet human expectations. Such formats include markdown for lists and pointers \citep{openai2024gpt4}, code blocks \citep{gur-etal-2023-understanding}, and integrate tags, or LaTeX for scientific texts \citep{singh2022progprompt, wang2023mathcoder}. Given the rising importance of formatting capabilities in LLMs, recently, format-following benchmarks have been developed for assessing LLMs' adherence to specified formats \citep{zhou2023instructionfollowing,xia-etal-2024-fofo, chen2024benchmarking,macedo2024exploring, 2024googmich}. However, these studies only evaluate format-instruction following capabilities. \emph{Our research further assesses LLM performance across different formats, uncovering significant format bias in various tasks and models}. We also acknowledge the concurrent work by \citet{tam2024let}, which examines the impact of format restrictions on LLM performance. However, unlike our approach, they do not disentangle evaluation metrics under format constraints and only evaluate $3$ structured formats, substantially fewer than our study.

\section{Output Format Evaluation Framework} \label{sec:format-eval-overview}
\subsection{Theoretical Analysis: Format Evaluation} \label{ssec:format-eval-metrics}


Automatic evaluation of LLMs {in question-answering and reasoning tasks} mainly relies on rule-based extraction to identify final answers from generated texts \cite{guo2023evaluating}. 
Within format constraints, determining the model's true performance, which is our focus, can be ambiguous and inaccurate, as correct responses might be overlooked due to format discrepancies (e.g., Case~(1) in  \Cref{fig:formateval-teaser}). To address this, we propose redefining these rule-based evaluation metrics to {reliably, transparently and accurately measuring the LLM performance given formats restrictions.} 

\paragraph{Notations.} Suppose that we are interested in evaluating an LLM $\mathcal{M}$ on a task $T$ using an evaluation metric $E$ (such as ``Accuracy'') under a format constraints $C$ (such as ``Bold the final answer.'') on $n$ samples with the ground-truth answers $\{y_1, ..., y_n\}$ and raw generated answers $\{\hat{y}_1, ..., \hat{y}_n\}$, where $y_i, \hat{y}_i \in \mathcal{Y}$ $\forall i$ with $\mathcal{Y}$ being the answer token sequence space. We denote $F_C$ as the binary format-following evaluation function of $C$:

\begin{equation}\label{eq:eq1}
F_C(\hat{y}_i)=
\begin{cases}
1, & \text{if}\ \hat{y}_i\ \text{satisfies}\ C. \\
0, & \text{otherwise.}
\end{cases}
\end{equation}

From \Cref{eq:eq1}, we define the {Format Instruction-following (FI) Score}, denoted as $FI_C$, as the percentage of generated outputs satsisfying $C$:

\begin{equation}\label{eq:eq2}
FI_C=\frac{\sum_{i=1}^{n} F_C(\hat{y}_i)}{n} \cdot 100
\end{equation}

Prior studies extensively focus on evaluating $FI_C$ \cite{zhou2023instructionfollowing,xia-etal-2024-fofo}. Our work further targets evaluating the \textbf{performance of LLMs given the format constraints} $C$. Under $C$, we denote $Ext_C()$ as the rule-based answer extractor (or a mixture of extractors) to extract the final answer from $\hat{y_i}$ for comparing it with $y_i$. We define: 
two evaluation scores based on $E$: 

\begin{definition}[\textbf{Systematic Evaluation Score ($SysE$)}]

\begin{equation} \label{eq:eq3}
SysE = \frac{1}{n}\sum_{i=1}^{n} (E(y_i,  Ext_C(\hat{y_i})) . F_C(\hat{y_i}))
\end{equation}
\end{definition}

Essentially, $SysE$ quantifies the performance of $\mathcal{M}$ on task $T$ based on the generated answers \emph{that meet the format constraints $C$}. {For example, in \Cref{fig:formateval-teaser}, Case~(1) yields a $SysE$ score of $0$, while Case~(3) achieves $1$}. This also shows that $SysE$ may not accurately reflect the actual performance of $\mathcal{M}$ on $T$, because $Ext_C()$ may fail to extract the final answers from (correct) answers 
dissatisfying $C$ {(e.g., Case~(1) in \Cref{fig:formateval-teaser})}. 
We define the {True Evaluation Score} to address this. Assume that we have 
{an oracle} extractor function $OracExt_C()$ that can extract the 
final answer from $\hat{y_i}$, regardless of whether $\hat{y_i}$ fulfills $C$, we have:

\begin{definition}[\textbf{True Evaluation Score ($TrueE$)}] \label{def:def2}

\begin{equation}\label{eq:eq4}
True E= \frac{1}{n}\sum_{i=1}^{n} E(y_i, OracExt_C(\hat{y_i}))
\end{equation}
\end{definition}

$TrueE$ measures the performance of $\mathcal{M}$ on task $T$ across all generated answers \emph{given the format constraints $C$, regardless of format satisfaction}. In \Cref{fig:formateval-teaser}, both Cases~(1)~and~(3) achieve a true accuracy of $1$. This score is crucial for assessing the true performance of LLMs given the format.

Prior studies do not clearly differentiate between $SysE$ and $TrueE$. In practice, measuring $TrueE$ is challenging because $OracExt_C()$ is unavailable. While researchers typically employ a mixture of methods to extract answers, this approach encounters two severe issues. First, these mixture-of-method extractors can be complex, unreliable, and often impractical for large-scale experiments with diverse formats like ours. Second, designing them to be reliable for complex formats such as medical reports can be impossible due to the countless potential errors. Another alternative is to assign a default value to $Ext_C(\hat{y_i})$. While this can temporarily avoid cases $\mathcal{M}$ fails to fulfill $C$, this is an incorrect practice since the default value may not be the actual output. Reliably measuring $TrueE$ often requires human investigation \citep{lin-etal-2022-truthfulqa} or the fine-tuning of evaluation models as scorers \citep{yang2024large}, both of which are costly. 

Nevertheless, $TrueE$ is crucial for a \emph{fair evaluation} of LLM performance bias across formats. Therefore, we propose a simple estimator of $TrueE$, denoted as $EstTrueE$: 

\begin{equation}\label{eq:eq5}
EstTrueE= 
\begin{cases}
SysE.\frac{100}{FI_C}, & \text{if}\ FI_C \neq 0. \\
  0, & \text{otherwise.}
\end{cases}
\end{equation}

{When $FI_C = 0$, estimating $EstTrueE$ becomes impossible}. $EstTrueE$ enables the fair format bias evaluation because normalizing $SysE$ by $FI_C$ prevents skewing comparisons of how different formats affect the LLM due to $FI_C$. It is especially useful for large-scale experiments since it is fully automatic. Let the $EstTrueE$ margin of error be $\epsilon$ with a confidence interval $1-\alpha$ and $S_C = n \cdot FI_C$ as \#generated answers satisfying $C$. 





\begin{theorem}[\textbf{Reliability of $EstTrueE$}] \label{theorem:theorem1}
$EstTrueE$ is consistent. Moreover, $EstTrueE$ is reliable if and only if:
\begin{equation} \label{eq:eq6}
FI_C \geq \frac{1}{1 + n \cdot \left( \frac{\epsilon}{ v \cdot s} \right)^2}
\end{equation}
\noindent Moreover, we have:
\vspace{3mm}
\begin{equation}
\lim_{{FI_C} \to 100} {EstTrueE} = {TrueE}
\end{equation}
\end{theorem}

\noindent where $s^2$ is the sample variance of evaluation scores of generated answers satisfying $C$ and $v = t_{\alpha/2, S_C-1}$ is the critical value from the t-distribution with $S_C-1$ degrees of freedom.




In summary, we have proposed a consistent estimator $EstTrueE$ of the true performance of LLMs measured by metric $E$ under the format constraints (Def.~\ref{def:def2}). This estimator is essential because it: (1) ensures transparent and fair LLM performance evaluation across different formats; and (2) supports large-scale format bias evaluation. Note that a high score $EstTrueE$ is only reliable iff $FI_C$ is high enough (\Cref{theorem:theorem1}). Henceforth, unless otherwise specified, $EstTrueE$ is our primary metric for measuring model performance given format constraints. The proof of \Cref{theorem:theorem1} is in \Cref{ssec:proofs}. 



\subsection{Theoretical Analysis: Format Bias} \label{ssec:format-eval-bias}

This section defines the metric to quantify format bias and outlines the criteria to mitigate such bias.

\paragraph{Bias measurement.} To measure the format bias of the LLM $\mathcal{M}$ across $k$ formats $F_o = \{C_1, \ldots, C_k\}$, we define a single metric, $BiasF_o$, as the variance of $EstTrueE$ scores over these $k$ formats, denoted as $\{EstTrueE_1, \ldots, EstTrueE_k\}$. Let $\mu_{EstTrueE} = \frac{1}{k} \sum_{i=1}^{k} EstTrueE_i$ represent the mean $EstTrueE$ score. Then:

\begin{equation} \label{eq:eq8}
BiasF_o = \frac{1}{k}\sum_{i=1}^{k}(EstTrueE_i - \mu_{EstTrueE})^2
\end{equation}

\paragraph{Realiability of $BiasF_o$.} By \Cref{eq:eq8}, the lower $BiasF_o$ is, the less format-$F_o$-biased $\mathcal{M}$ is, suggesting a criterion for mitigating output format bias. However, $BiasF_o$ is an estimator based on the estimators $EstTrueE_i$. Therefore, to enhance the reliability of $BiasF_o$, it is also necessary to improve the reliability of $EstTrueE_i$ by increasing $FI_{C_i}$ $\forall i$ (\Cref{theorem:theorem1}). Therefore, we propose \textbf{two necessary criteria for an effective method to mitigate format bias in LLMs}: \textbf{(i) Minimize bias metric:} reducing $BiasF_o$, indicating less format-$F_o$-bias in $\mathcal{M}$; \textbf{(ii) Increase the format-following scores for all formats:} ensuring the reliability of $BiasF_o$ by increasing the $FI$ scores across all the formats: $\{FI_{C_1}, ..., FI_{C_k}\}$ (\Cref{eq:eq2}).


\subsection{Formats for Evaluation} \label{ssec:format-eval-categories}


We establish $4$ format categories for evaluation consisting of $15$ formats introduced by prior practices:

\paragraph{(i) Multiple-choice question (MCQ) answer (\Cref{sec:mcq}).} where LLMs answer questions by selecting from provided choices, presented as either a \textbf{(1) Character identifier} \cite{robinson2023leveraging}; or \textbf{(2) Choice value} \cite{chen2023universal}. 

\paragraph{(ii) Wrapping (\Cref{sec:wrapping}).} where LLMs must enclose the final answer within the two characters, which is crucial for automatic evaluation to isolate the final answer from reasoning thoughts. We focus on evaluating $7$ widely used wrapping strategies: 
\textbf{(1) Special character} \cite{gur-etal-2023-understanding}; \textbf{(2) Bolding} \citep{zhou2023instructionfollowing}; \textbf{(3) Italicizing} \citep{zhou2023instructionfollowing}; \textbf{(4) Double brackets} \citep{luo2024large}; \textbf{(5) Double parentheses}; \textbf{(6) Placeholder} \citep{wang2023mathcoder}; \textbf{(7) Quoting} \citep{zhou2023instructionfollowing}. 

\paragraph{(iii) List (\Cref{sec:list}).} where the output of LLMs is a list of elements. We investigate $4$ formats representing lists: \textbf{(1) Python list} \cite{do2023choire}; \textbf{(2) Bullet-point list} \cite{2024googmich}; \textbf{(3) List of elements separated by a special character ``[SEP]''} \cite{Boucher_2023}; and \textbf{(4) List of elements arranged on separate lines} \cite{Mishra_2023}.

\paragraph{(iv) Mapping (\Cref{sec:mapping}).} where LLMs are employed to output dictionaries or maps. We focus on two ubiquitously used mapping structures: \textbf{(1) Python dictionary/JSON} (JavaScript Object Notation) \citep{baumann2024using} and \textbf{(2) YAML} (Yet Another Markup Language) \citep{goel2023llms}. 




\paragraph{Format-instruction following.} We introduce Appx.-\Cref{algo:fi-score}, a rule-based heuristic to determine the format-instruction following function $F_C$ (\Cref{eq:eq1}) for our benchmarked formats. It calculates the binary FI score by verifying that the generated output includes the specified formatting tokens and that the extracted final answer matches the expected type. It is highly extendable to other formats (\Cref{appdx:fi-scorer}).

\section{General Experimental Setups} \label{sec:general-setups}



\paragraph{Benchmarks.} For MCQ bias evaluation (\Cref{sec:mcq}), we select two datasets: \textbf{MMLU} \citep{hendrycks2020measuring} and \textbf{BBH} \citep{suzgun-etal-2023-challenging}. For MMLU, we randomly choose 27 subcategories. For BBH, we select the \texttt{sports\_understanding} category following \citet{gupta2024bias}. For wrapping bias assessment (\Cref{sec:wrapping}), in addition to MCQ benchmarks, the following datasets are experimented: \textbf{GSM8K} \citep{cobbe2021gsm8k} for reasoning, \textbf{FairytaleQA} \citep{xu-etal-2022-fantastic} for narrative comprehension, and \textbf{HotpotQA} \citep{yang-etal-2018-hotpotqa} for multi-hop reasoning. For list bias investigation (\Cref{sec:list}), we use \textbf{SciDocsRR} \citep{muennighoff-etal-2023-mteb}, a scientific document ranking task as the order list generation task, and \textbf{SemEval 2017} \citep{augenstein-etal-2017-semeval}, the keyphrase extraction task as the unordered list generation. For mapping bias examination (\Cref{sec:mapping}), we utilize a document-level information extraction task named \textbf{SciREX} \citep{jain-etal-2020-scirex} by synthesizing three extraction difficulty levels: easy (extracting from $1$ sentence for $1$ category), medium ($3$ sentences, $2$ categories), and hard ($5$ sentences, $4$ categories). For all benchmarks except MCQ, we sample $200$ points for evaluation \cite{bai2023longbench}.

\paragraph{Models.} We select both open- and closed-source LLMs for our evaluation: \textbf{Gemma-7B-it} \citep{team2024gemma}, \textbf{Mistral-7B-it-v0.2} \citep{jiang2023mistral}, {and \textbf{Llama-3.1-8B-it} \citep{dubey2024thellama3}} for open-source as they are among state-of-the-art open-source LLMs; \textbf{ChatGPT (gpt-3.5-turbo-0125)} for closed-source as this premier chatbot possesses superior instruction-following ability. Our purpose is not to reproduce the models' performance, but to show the bias. 

\paragraph{Metrics.} Following our discussion in \Cref{ssec:format-eval-metrics}, we disentangle \textbf{Accuracy (Acc)} for MMLU and BBH \citep{guo2023evaluating}; \textbf{F1} for GSM8K, HotpotQA, FairytaleQA; and \textbf{Mean Average Precision (MAP)} for SciDocsRR \citep{muennighoff-etal-2023-mteb} and we report the metrics $EstTrueAcc$, $EstTrueF1$, $EstTrueMAP$ (\Cref{eq:eq5}) in the main text. For metrics' reliability, we set $\alpha$ = $\epsilon=5\%$.


\paragraph{Prompting baselines.} Our focus is on two widely used prompting baselines: (1) \textbf{Zero-shot (ZS)} prompting and (2) \textbf{Zero-shot Chain-of-Thought (ZS-CoT)} prompting \citep{kojima2022large}. For the ZS baseline, we instruct LLMs to answer the question with the prompt ``Answer the following question...'' followed by the suffix ``without any explanation''. 
For ZS-CoT, we use the suffix ``step-by-step'' instead. For the ZS-CoT experiments in \Cref{sec:mcq,sec:list,sec:mapping}, LLMs are instructed to wrap the final answer by ``<ANSWER>'' and ``</ANSWER>'' tokens to distinctly isolate it from the reasoning chains (see \Cref{tab:wrapping-tokens} for the wrapping instruction). We use this wrapping method since our experiment in \Cref{sec:wrapping} shows that it achieves the highest instruction-following score on average across LLMs. Detailed prompts are provided in \Cref{appdx:prompting}. We average the performance under two prompting methods to report in the main text. 


\section{Format Evaluation Experiments}

Overall, we find that: (1) Models show substantial format-following bias across formats for all benchmarks; (2) For all models and datasets, significant performance bias exists across formats; (3) {$78.30\%$} of the $EstTrue$ results are reliable (with {$70\%$} for MCQ, {$82.5\%$} for wrapping, {$67.19\%$} for list, and {$77.08\%$} for mapping formats) \emph{highlighting significant weaknesses of LLMs in following format instructions}. We dive into (2) for every format as it is our main focus, (1, 3) are discussed in detail in \Cref{appdx:mcq-discussions,appdx:wrapping-discussions,appdx:list-discussions,appdx:mapping-discussions}.

\subsection{Experiments on MCQ Format} \label{sec:mcq}





\paragraph{Setups.} \label{sec:mcq-setups}
We investigate the bias of LLMs towards different MCQ output formats. We assess two formats as introduced in \Cref{ssec:format-eval-categories}: (1) Character identifier and (2) Choice value. For example, if the choice is ``[A. Yes, B. No]'', then the character identifier can be ``A/B'', while the choice value can be ``Yes/No''. We exclude the format combining the character identifier and choice value (such as ``A. Yes'') from our evaluation because instructing LLMs to output this format can be non-trivial and require manual effort to craft instructions tailored for different models. To ensure that LLMs understand the ``Character identifier'' and ``Choice value'' as we expect, we add a contrastive format requirement to the prompts (e.g., ``without any textual description'' for the ``Character identifier'' prompts). 

\begin{figure}[!tp]
\centering
\includegraphics[width=1\linewidth]{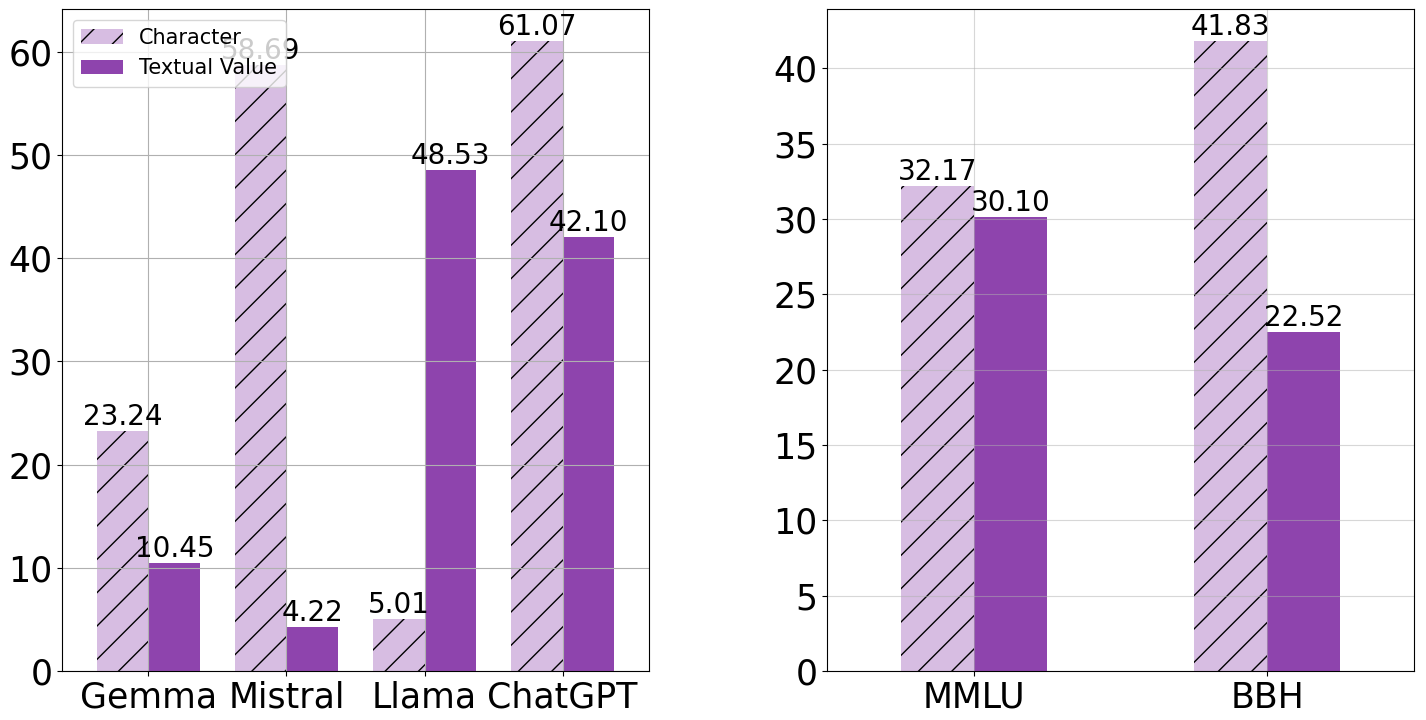}
\caption{Average estimated true accuracy (\Cref{ssec:format-eval-metrics}) results of MCQ benchmarks across models (left) and datasets (right) showing performance bias of LLMs across formats.
}
\label{fig:mcq_main}
\end{figure}

\paragraph{Results.} \label{sec:mcq-findings}
\Cref{fig:mcq_main} provides a synopsis of our evaluation results, with numerical values shown in Appx.-\Cref{tab:mcq-results}. From \Cref{fig:mcq_main}-left, we observe that Mistral possesses the highest disparity between the two MCQ answer formats, with {$58.69\%$} accuracy on average for character and only {$4.22\%$} for textual value. Additionally, despite ChatGPT often being regarded as one of the most robust LLMs, it shows a significant performance difference between the two formats ({$19.03\%$}). Overall, LLMs are heavily biased towards outputting character identifiers. Requiring them to generate the choice's value causes notable performance drops on most models.

From \Cref{fig:mcq_main}-right, we notice that the models exhibit higher bias on BBH, which appears to be an easier benchmark than MMLU. We attribute this to the small size of BBH, which makes the performance more sensitive to format variations.


\paragraph{Why such bias?} 
We hypothesize the root cause of the significant performance bias across different formats is the \textbf{format token bias} of LLMs. The non-uniform distribution of FI scores among formats suggests that the models assign probabilities to format instructions differently based on their training data. This leads to varying prior assignments of probabilities to specific tokens, causing final predictions non-uniformly distributed across formats. This hypothesis is supported by our simple fine-tuning with formatted data, which familiarizes LLMs with format instructions relatively equally leading to a drastic format bias reduction (\Cref{sec:mitigating-format-bias}). This emphasizes the necessity of more research in fine-tuning LLMs to reduce format bias and raises concerns about the reliability and reproducibility of recent studies using varied formats.



\subsection{Experiments on Wrapping Format} \label{sec:wrapping}

\begin{table}[!tp]
\small 
\resizebox{1\linewidth}{!}
{
\begin{tabular}{l|c|l}
\toprule
\textbf{Wrapping type} & \textbf{(start, end)} & \textbf{Prompt:} Wrap your final answer... \\ 
\midrule
\midrule
Special char. & (<ANSWER>, </ANSWER>) & by <ANSWER> and </ANSWER>. \\
\midrule
Bolding & (**, **) & in bold by enclosing it with double asterisks.\\
\midrule
Italicizing & (*, *) & in italics by enclosing it with single asterisks.\\
\midrule
Brackets & ([[, ]]) & using double square brackets. \\
\midrule
Parentheses & (((, ))) & using double parentheses.\\
\midrule
Placeholder & None & by filling in the placeholder below: \\
 & & ``So the answer is: [placeholder]'' \\
\midrule
Quoting & ('''''', '''''') & using triple double-quotation marks.\\
\bottomrule
\end{tabular}
}
\caption{\small{Wrapping ``start'' and ``end'' tokens with instructions.}}
\label{tab:wrapping-tokens}
\end{table}

\begin{figure*}
\centering
\subfloat{\includegraphics[width=0.49\linewidth]{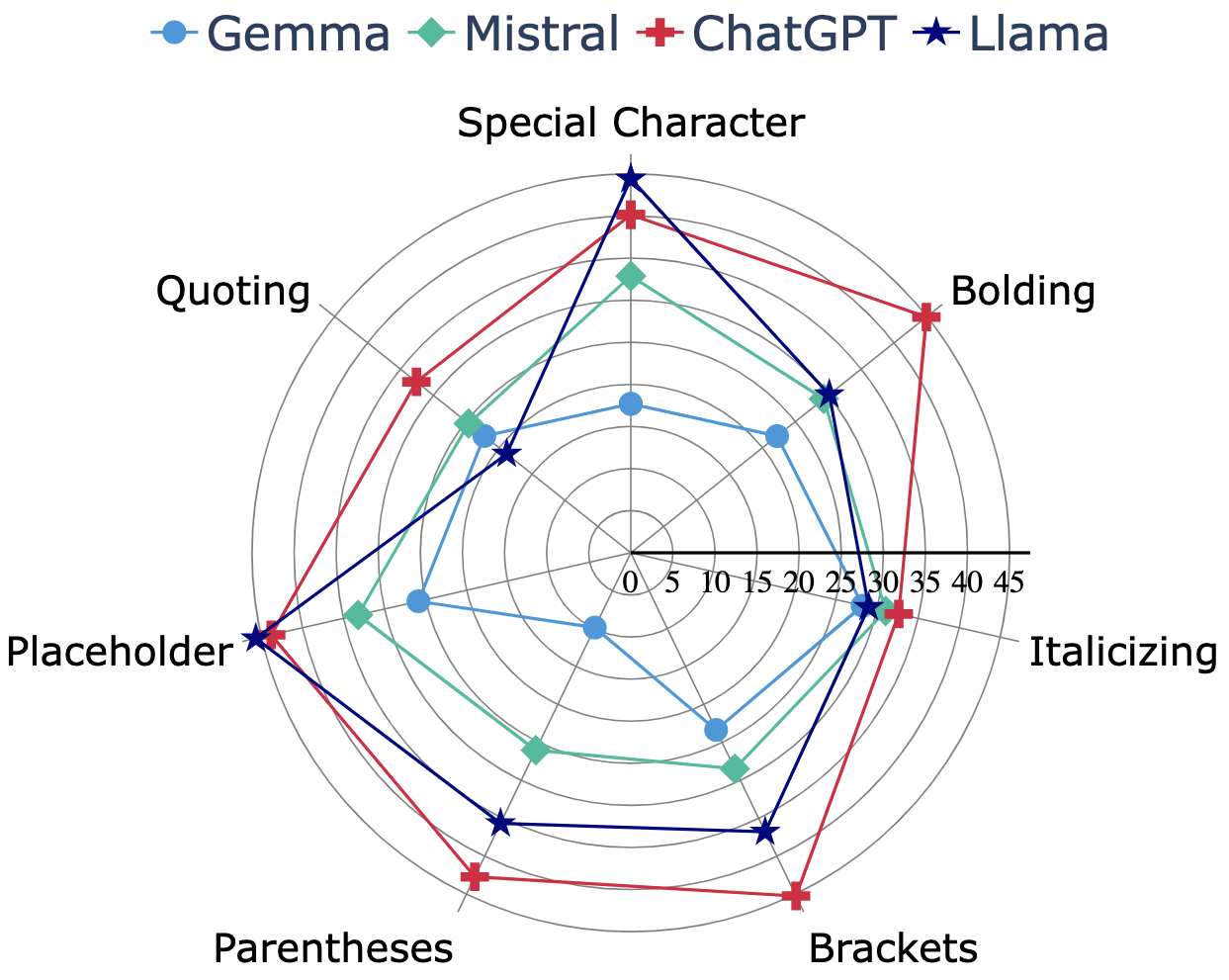}}
\subfloat{\includegraphics[width=0.48\linewidth]{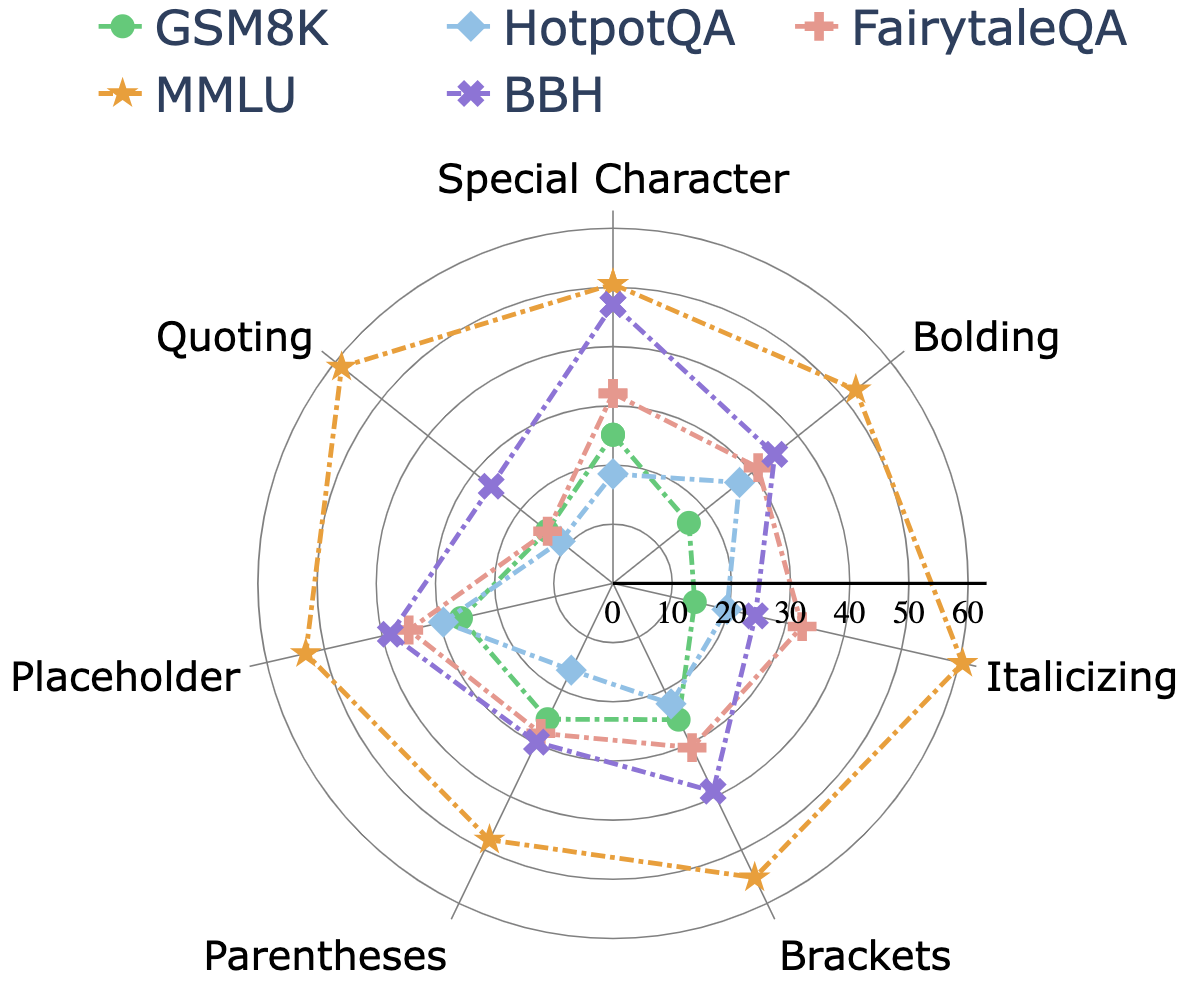}}
\caption{\small{Average estimated true Accuracy (MCQ) and F1 (GSM8K, HotpotQA, FairytaleQA) scores (\Cref{ssec:format-eval-metrics})  across models (left) and across benchmarks (right), showing performance bias of LLMs across $7$ widely used wrapping methods. 
}} 
\label{fig:wrapping-bias}
\end{figure*}

\paragraph{Setups.} \label{ssec:wrapping-setups}
We study LLM bias towards $7$ wrapping methods: (1) Special character; (2) Bolding; (3) Italicizing; (4) Brackets; (5) Parentheses; (6) Placeholder; (7) Quoting, detailed in \Cref{tab:wrapping-tokens}. We evaluate LLM performance across formats on the MMLU, BBH, GSM8K, FairytaleQA, and HotpotQA. 

\paragraph{Results.} \label{ssec:wrapping-findings} 

\Cref{fig:wrapping-bias} outlines an overview of our evaluation outcomes with  results in Appx.-\Cref{tab:wrapping-results}. 
From \Cref{fig:wrapping-bias}-left, we see that Llama exhibits the highest bias towards different formats (with a $BiasF_o$ value of {$74.86\%^2$}; see Appx.-\Cref{Wrapping:values}), while ChatGPT performs the best. Notably, for ``Quoting'' and ``Parenthesis'', the Gemma follows instructions only about $0-4\%$ yielding nearly zero performance, highlighting its critical weaknesses. Among the 7 formats, ``Placeholder'' ({$37.15\%$}) proves to be the most effective wrapping output format, while ``Quoting'' ({$24.58\%$}), ``Parenthesis'' ({$28.57\%$}) are among those that achieve the lowest performance.

From \Cref{fig:wrapping-bias}-right, models exhibit bias across all tasks, with the lowest on MMLU ($16.58\%^2$; see Appx.-\Cref{Wrapping:values}) possibly because the models already performed relatively well on it, and the highest on BBH ($54.58\%^2$), the challenging task without train data. This demonstrates the pervasive presence of wrapping bias in LLMs.

\paragraph{Why such bias?} The format token bias of LLMs as explained in \Cref{sec:mcq} is also our hypothesis. Specifically, we found the low performance of the ``Quoting'' and ``Parenthesis'' because, in generation tasks, models often wrap (via quoting/parenthesizing) not only the final answer, as instructed, but also parts of the context (e.g., ```The answer is 3.'''), leading to poor F1 scores. Moreover, Gemma completely ignores the above format instructions, resulting in 0\% FI scores, which also contribute to the low average estimated F1 scores. These strongly indicate the presence of format token bias.

\subsection{Experiments on List Format} \label{sec:list}

\paragraph{Setups.}
We explore the bias of LLMs in generating lists following $4$ formats: (1) Python list, (2) Bullet-point list, (3) Character-separated list, and (4) Newline-separated list. We evaluate the models on two list generation tasks: \emph{(i) Unordered list}, using the keyphrase extraction task on the SemEval 2017 dataset, and \emph{(ii) Ordered list}, using the document ranking problem on the SciDocsRR task. 

\begin{figure}[!htp]
\centering
\subfloat{\includegraphics[width=0.5\linewidth]{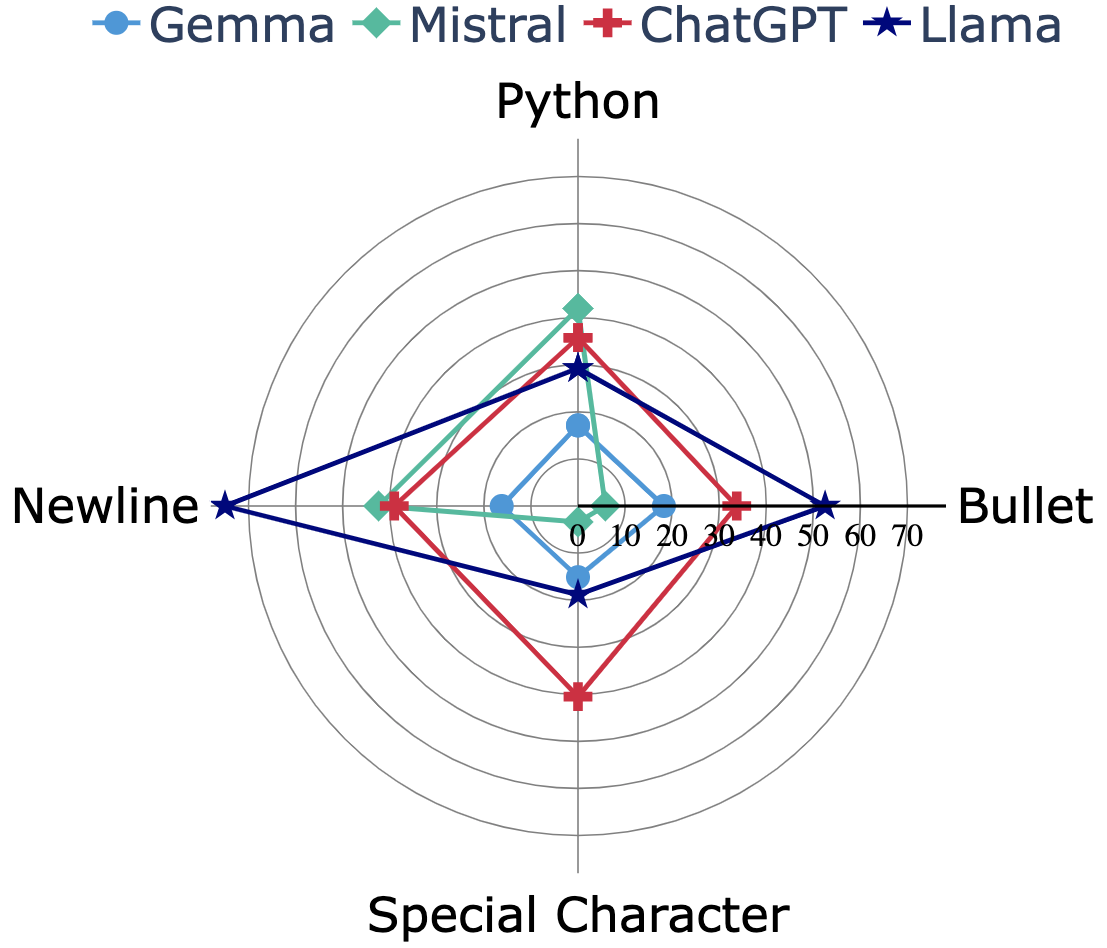}}
\subfloat{\includegraphics[width=0.5\linewidth]{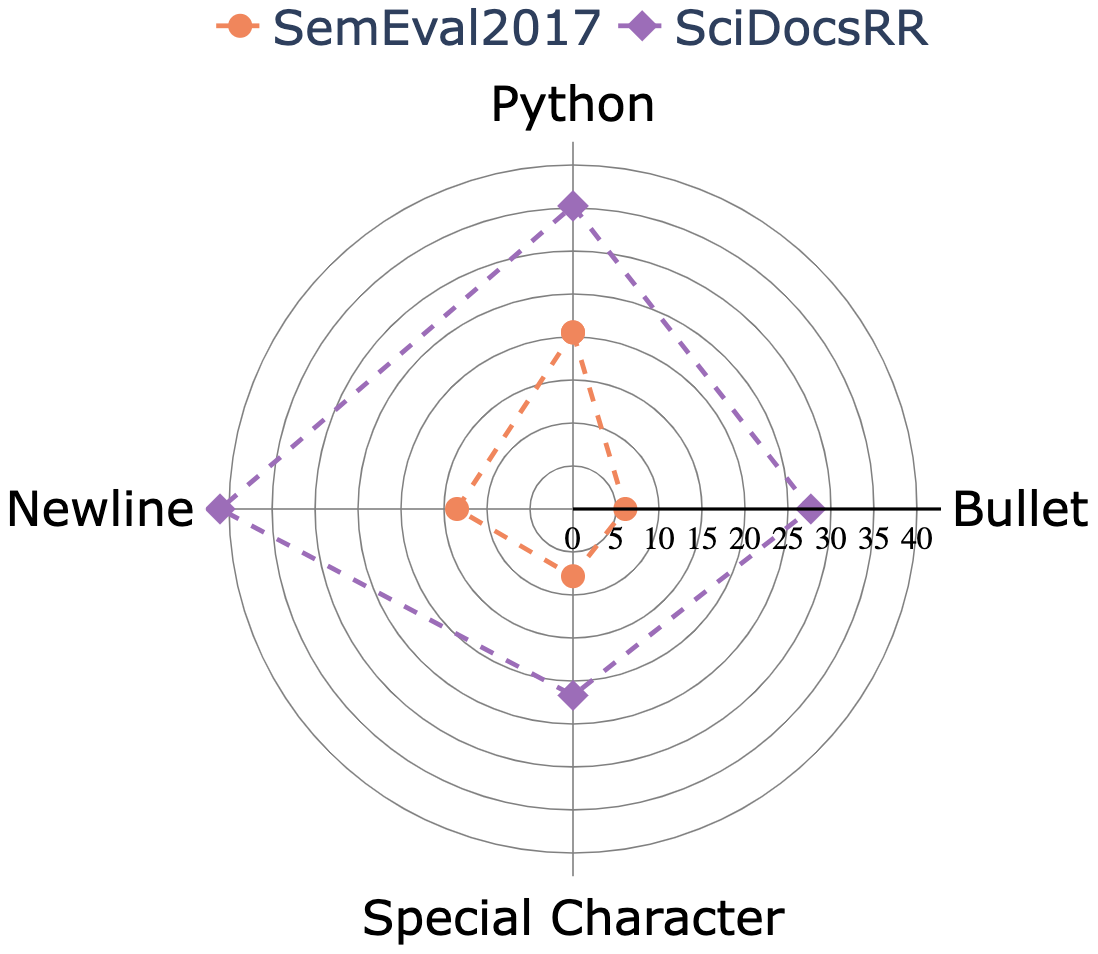}}
\caption{\small{Average $EstTrueF1$ (SemEval2017) and $EstTrueMAP$ (SciDocsRR) (\Cref{ssec:format-eval-metrics}) across models (left) and benchmarks (right) showing performance difference of LLMs across $4$ widely used list formats.
}}
\label{fig:list-bias}
\end{figure}

\paragraph{Results.} 
\Cref{fig:list-bias} displays the key findings of our evaluation across models and datasets with numerical results in Appx.-\Cref{tab:list-results}. From \Cref{fig:list-bias}-left, we notice that Mistral exhibits the most bias, with the $BiasF_o$ value of $353.80\%^2$. In contrast, ChatGPT and Gemma show much lower bias, with values of $7.08\%^2$ and $1.32\%^2$, respectively. Of the four formats, the ``Python'' and ``Newline-separated'' formats yield the highest performance, likely due to models trained extensively on code data. Conversely, the ``Bullet-point list'' format results in the lowest performance, particularly for Mistral, highlighting the inherent bias for such formats.


The performance bias is regardless of the task as plotted in \Cref{fig:list-bias}-right, with the highest $BiasF_o$ value of {$54.12\%^2$} on the order list generation task SciDocsRR, and significantly lower ({$31.86\%^2$}) on SemEval2017 task. The high bias in the SciDocsRR task is because Mistral and Gemma mostly failed to perform this task following the ``Bullet'' and ``Special character'' list formats while excelling in solving it following the other formats.

\paragraph{Why such bias?}
We attribute the bias to the format token bias (\Cref{sec:mcq}). Since the models were extensively trained on code data, they excel in solving code-related instructions. In contrast, ``Bullet-point'' and ``Special character'' lists are much less common. One interesting case is Gemma where it performed worse on generating ``Python'' lists compared to ``Bullet-point'' lists. Our analysis suggests that Gemma misinterprets the format instruction as a coding request, generating Python code programs instead of an answer in a Python list, suggesting Gemma was predominantly trained on code data.

\subsection{Experiments on Mapping Format} \label{sec:mapping}

\paragraph{Setups.}

We examine the performance bias of LLMs on two mapping formats as discussed in \Cref{sec:format-eval-overview}: (1) Python dictionary/JSON; (2) YAML. We preprocess the SciREX task \cite{jain-etal-2020-scirex} as described in \Cref{sec:general-setups} into three extraction levels: (i) Easy (1 sentence, ``Task'' category); (2) Medium (3 sentences, ``Task, Method''); (3) Hard (5 sentences, ``Task, Method, Material, Metric'' categories). 

\paragraph{Results.} \label{ssec:mapping-findings}

\begin{figure}[!tp]
\centering
\includegraphics[width=1\linewidth]{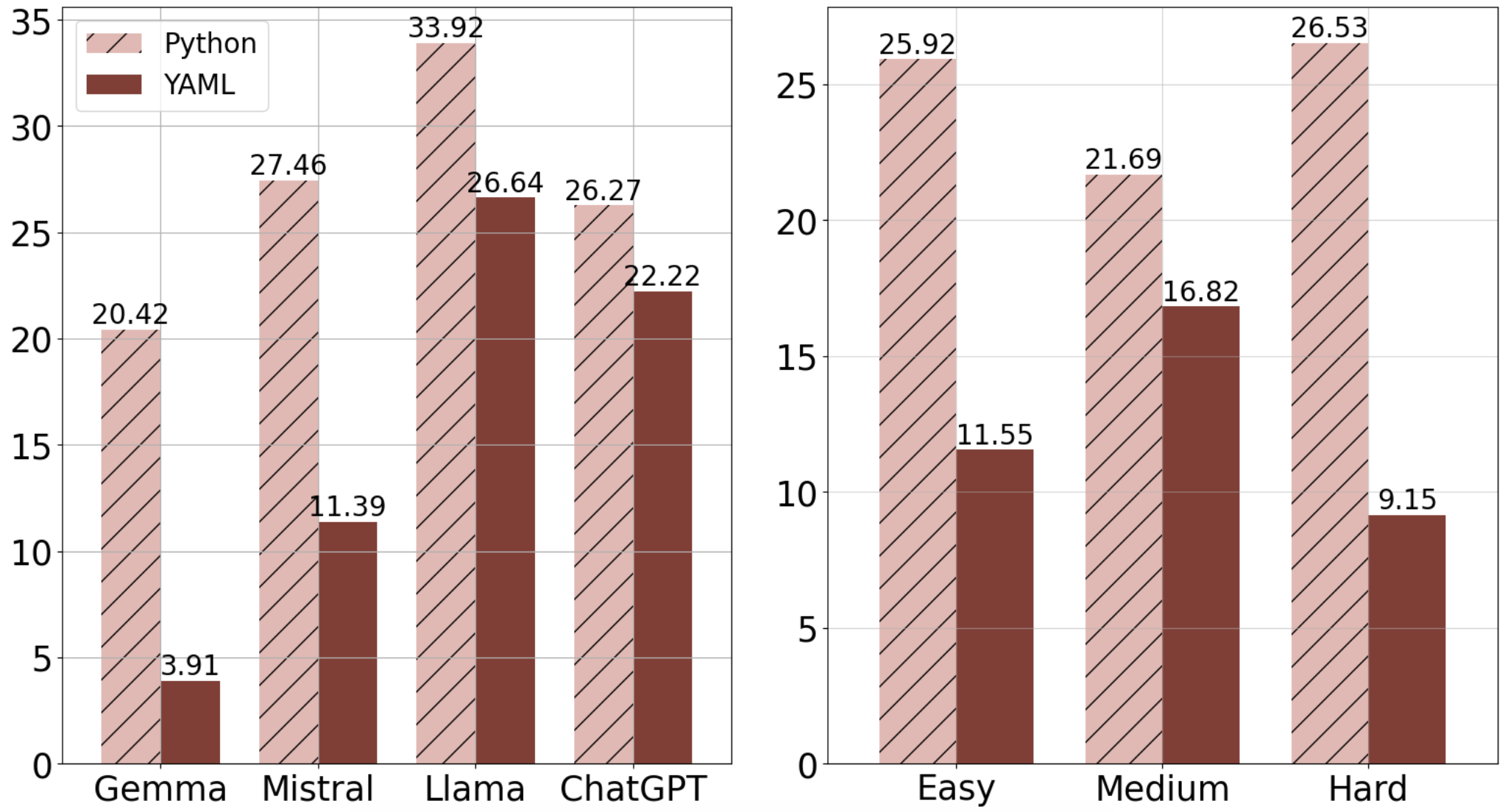}
\caption{\small{Average estimated true F1 scores (\Cref{ssec:format-eval-metrics}) across models (left) and benchmarks (right) showing performance bias of LLMs across $2$ widely used mapping formats. 
}}
\label{fig:mapping_main}
\end{figure}

\Cref{fig:mapping_main} illustrates a summary of our evaluation with numerical details in Appdx.-\Cref{tab:mapping-results}. From \Cref{fig:mapping_main}-left, Gemma is the most biased, with a performance gap of $16.51\%$ between the two formats, followed by Mistral with a $16.07\%$ gap. ChatGPT and Llama, however, are relatively robust against format variations. On average, JSON performs significantly better than YAML for mapping, likely because more JSON data is used to train models due to its popularity.



From \Cref{fig:mapping_main}-right, extracting $4$ categories in the Hard task shows the largest performance gap between mapping formats. Surprisingly, the Medium task displays the least bias, likely because models perform best in this task.

\paragraph{Why such bias?} 
The bias is attributed to the format token bias (\Cref{sec:mcq}). While Mistral excels in generating JSON, it and Gemma struggle with YAML. Even successfully generating YAML output, Mistral and Gemma frequently introduce noisy information (88\%-65\% for Mistral with and without CoT, 98\%-79\% for Gemma) in the response (e.g., a key ``Task" should have multiple values, Mistral generates multiple key-value pairs instead e.g., ``Task\_1:Training $\cdots$ Task\_2: $\cdots$''), resulting in poor overall performance.





\section{Mitigating Performance Format Bias: Actionable Recommendations} \label{sec:mitigating-format-bias}

\begin{figure}[!tp]
\centering
\includegraphics[width=1\linewidth]{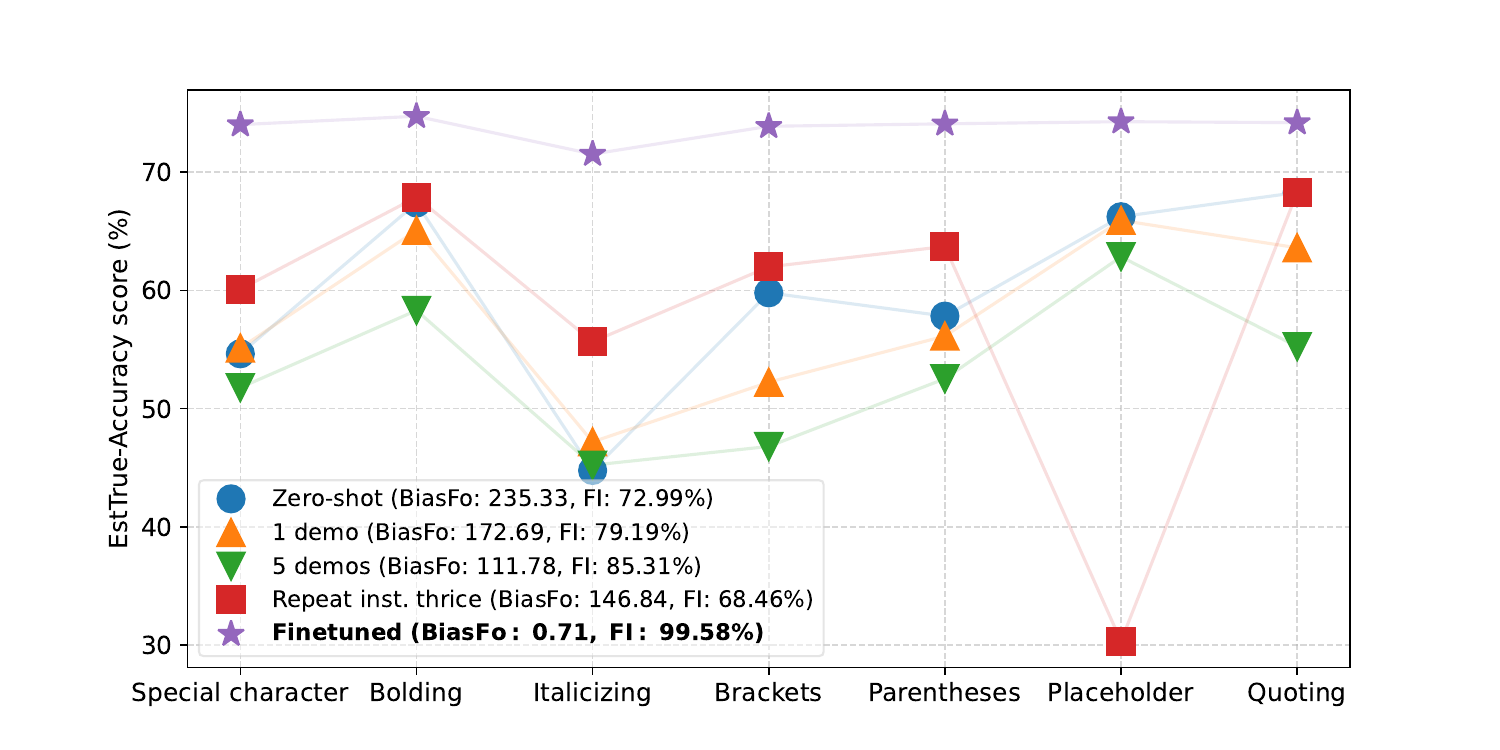}
\caption{\small{More demonstrations and repeating format instructions mitigate format bias. Finetuning mostly eliminates the format bias. The performance is reported using ChatGPT on MMLU (Appx.-\Cref{tab:more-demonstrations} for num. results).}}
\label{fig:more-demos-help}
\end{figure}

We propose methods as actionable recommendations to mitigate format bias in the performance of LLMs. Generally, three primary streams of techniques have been widely studied and applied to tackle LM biases: (1) Prompting \cite{xu2024take,macedo2024exploring}; (2) Calibrating \cite{roelofs2022mitigating,li2024mitigating}; and (3) Fine-tuning \cite{schick-etal-2021-self,ghaddar-etal-2021-end}. While calibration techniques can only be used for white-box models, prompting and fine-tuning can be applied for both black-box (via API) and white-box ones. Therefore, we explore prompting and fine-tuning techniques to reduce format bias. We target mitigating the format bias of \textbf{ChatGPT} in \Cref{fig:more-demos-help}, the strongest model that we benchmarked, on \textbf{MMLU}. We aim to reduce the \textbf{wrapping} bias (\Cref{sec:wrapping}) due to resource limits, but our methods can be generalized to any model and format. We also verify (and confirm) our mitigation strategies on \textbf{Gemma-2B-it}, a medium-size model in Appx.-\Cref{tab:more-demonstrations}.

\paragraph{Demonstration(s) reduce(s) format bias.} \label{ssec:demonstrations-help-reduce-bias}

As discussed in \Cref{sec:mcq-findings}, LLMs show bias across formats possibly because of the token bias issue, causing LLMs to non-uniformly comprehend the format instructions. To address this, we examine whether demonstrations with formats can reduce such bias, as they are commonly utilized to enhance LLM's comprehension of the task patterns \cite{xie2022an}. Particularly, for each wrapping format in \Cref{sec:wrapping}, we select $1$ and $5$ random samples from the auxiliary train data of MMLU and manually format the answers as demonstrations. The results are outlined in \Cref{fig:more-demos-help}. Firstly, incorporating demonstrations typically enhances the FI scores of the model (from 72.99\% to 79.19\% and 85.31\%) (i), with five demonstrations yielding the most. Secondly, we observe a notable decrease in the $BiasF_o$ score (ii) upon supplementing demonstrations. From (i), (ii) and \Cref{ssec:format-eval-bias}, we conclude integrating demonstrations mitigates format bias.






\paragraph{Repeating format instructions reduces format bias.} 
We found that repeating instructions generally increases FI scores (i) across most formats except ``Placeholder'', which can consequently lessen the mode's token bias towards format instructions (\Cref{sec:mcq}). Using our two proposed criteria for effective format bias mitigation in \Cref{ssec:format-eval-bias}, it is worth examining if this approach reduces $BiasF_o$, thereby being an effective mitigation. Our answer is yes. By repeating the wrapping instructions of ChatGPT thrice, we observed a decrease in the $BiasF_o$ (ii) score presented in \Cref{fig:more-demos-help}. Combining (i) and (ii) suggests that this strategy is an effective mitigation. For "Placeholder," human investigation reveals that multiple placeholder instructions cause ChatGPT to be confused about where the placeholder is, making it frequently misunderstand and fail to follow this format instruction.


\paragraph{Fine-tuning with additional format data can eliminate format bias.}
We hypothesize that completely solving the format token bias problem of LLMs necessitates finetuning them on format data so that they are familiar with tokens in format instructions evenly. We propose a simple data synthesis strategy for finetuning LLMs: we {sample a small set of training data for all evaluated formats, with ratios inversely proportional to their \textbf{systematic evaluation scores} (\Cref{ssec:format-eval-metrics})}. We chose $SysE$ scores over the $EstTrueE$ because they reflect the current model performance. Practically, based on ChatGPT's zero-shot systematic performance on MMLU colored in \textcolor{blue}{blue} in Appx.-\Cref{tab:wrapping-results}, we approximate the formats' performance ratios as ``$1,1,\frac{1}{2},\frac{1}{2},\frac{1}{3},1,\frac{1}{3}$'' from left-to-right, resulting in training data ratios of formats of ``$1,1,2,2,3,1,3$''. We then preprocess the MMLU auxiliary training data according to these ratios, scaled by $500$ ($6500$ samples total), and train ChatGPT on this dataset. 
The finetuned results are plotted in \Cref{fig:more-demos-help}. Firstly, after finetuning, the average FI score across all formats is nearly perfect at $99.58\%$ (ii). Secondly, the $BiasF_o$ score is significantly reduced from $235.33\%^2$ to $0.71\%^2$ (ii). These (i) and (ii) indicate finetuning largely eliminates format bias.

\section{Conclusions}

We introduce the pioneering systematic investigation of format bias in LLM performance, revealing significant biases across widely used formats for all models and benchmarks. Our method involves developing metrics to assess this bias and establishing criteria for effective mitigation. We then introduce prompting and fine-tuning techniques to alleviate format bias based on our evaluation findings. Our work aims to sharpen the focus of future LLM research toward fairer and more robust development.


\section*{Limitations}
Our study has several limitations. Firstly, the metrics $EstTrue$ and $BiasF_o$ proposed in \Cref{ssec:format-eval-metrics} and \Cref{ssec:format-eval-bias} are estimators, not exact measures. As discussed, determining $TrueE$ (\Cref{eq:eq4}) is infeasible, especially for large-scale experiments across various models and datasets. Achieving this would require extensive fine-tuning and comprehensive human evaluations, both prohibitively expensive and impractical in many scenarios. Our proposed metrics $EstTrue$ and $BiasF_o$ are handy for large-scale experiments with multiple models and datasets due to their fully automatic nature. We further propose \Cref{theorem:theorem1} to validate the reliability of $TrueE$ statistically. 

Secondly, our empirical evaluation of format bias is limited by computational and budget constraints to specific datasets, formats, and models. This restriction limits the generalizability of our findings and may obscure further insights that could be gained from expanding the experiments to include more formats, larger-scale datasets, and additional task categories.

Finally, while our study primarily attributes format bias to token bias in the training data of LLMs and proposes data-focused approach, it does not extensively explore other factors related to model architecture and training processes. This omission represents a significant area for future research, as more fundamental, architecture-level solutions could be crucial, for addressing format bias in LLMs. Our study underscores the importance of continued research dedicated to quantifying and mitigating format bias.

\section*{Ethical Considerations}
Our work uncovers significant format bias in LLMs, raising concerns regarding fairness and potential discrimination in real-world applications.

\paragraph{Bias and fairness.} Format bias in LLMs can result in unfair treatment, especially in tasks where multiple possible formats can be used. Our research suggests ways to identify and mitigate format bias, aiming for fairer and more equitable LLM applications.

\paragraph{Societal impact.} 
Format bias in LLMs has the potential to disproportionately impact specific populations, as different demographics may have preferences for different communication formats. Further research is essential to fully understand its societal implications and ensure fairness across diverse demographics.

\section*{Acknowledgements}
This research project is partially supported by the National Research Foundation Singapore under the AI Singapore Programme (AISG Award No: AISG2-TC-2023-010-SGIL), the Singapore Ministry of Education Academic Research Fund Tier 1 (Award No: T1 251RES2207), and the National Research Foundation, Singapore under its AI Singapore Programme (AISG Award No: AISG2-GC-2022-005). DXL and TS are supported by the A*STAR Computing and Information Science (ACIS) scholarship. We thank Goh Yisheng for his contribution in the initial stage of the project. We thank members of WING and Deep Learning Lab, NUS, as well as the anonymous reviewers for the constructive feedback.

\bibliography{anthology,custom}
\bibliographystyle{acl_natbib}

\onecolumn
\appendix
\newpage

\section{Format-Instruction Following Scorer} \label{appdx:fi-scorer}

\begin{algorithm*}[ht]
\renewcommand{\algorithmicrequire}{\textbf{Input:}}
\renewcommand{\algorithmicensure}{\textbf{Output:}}
\caption{Format-Instruction Following Scorer}
\begin{algorithmic}[1]
\REQUIRE Task $T$, language model $\mathcal{M}$, format constraints $C$, generated output $Y$.
\REQUIRE If $C$ includes wrapping characters, we denote as $\{W_1, W_2\}$ and $is\_wrapping = True$.
\REQUIRE $output\_type$ is the data type required by $C$ when $T$ is not MCQ.

\IF{$is\_wrapping$}
    \RETURN False if (any of $\{W_1, W_2\} \not\in Y$) or (number of $W_1 \in Y$ + number of $W_2 \in Y$ $\neq$ 2).
    \STATE $ans$ = Extract string in between $\{W_1, W_2\}$.
\ELSE
    \STATE $ans = Y$
\ENDIF

\IF{$T$ is MCQ}
    \IF{MCQ output type is character identifier}
        \RETURN True if $ans \in \{A, B, C, D\}$. False otherwise.
    \ELSE
        \RETURN True if $ans \in$ \{options' values\}. False otherwise.
    \ENDIF
\ELSE
    \RETURN True if we can parse $ans$ as an instance of the class $output\_type$. False otherwise.
\ENDIF    
\end{algorithmic}
\label{algo:fi-score}
\end{algorithm*}

\Cref{algo:fi-score} presents our heuristic algorithm for evaluating the format-instruction following capabilities of LLMs, which is used to compute $F_C$ in \Cref{eq:eq1}. The algorithm is divided into two three main parts:

\begin{enumerate}
    \item \textbf{Lines 1-6}. These lines focus on examining the wrapping requirements by verifying the presence and correctness of the specified wrapping tokens.
    \item \textbf{Lines 7-12}. These lines are dedicated to checking the formats of MCQ answers (\Cref{sec:mcq}).
    \item \textbf{Lines 13-15}. These lines address the remaining formats, including list and mapping formats.
\end{enumerate}

It is worth noting that \Cref{algo:fi-score} is highly adaptable; formats can be added or removed to tailor it for specific downstream applications.

\section{Theoretical Analysis: Reliability of $EstTrueE$} \label{sec:reliability-of-esttruee}

\subsection{Proof of \Cref{theorem:theorem1}} \label{ssec:proofs}

\begin{proof}[\textbf{Proof of \Cref{theorem:theorem1}}]

We omit the case when $FI_C=0$ since in that case, we cannot estimate $TrueE$. By the definition in \Cref{theorem:theorem1}, we have $S_C$ generated answers that satisfy $C$. Let's denote $k=S_C$ for simplicity. Let's denote $k$ performance scores of answers satisfying $C$ as $x_1,\cdots,x_k$, and $\Bar{x} = \frac{\sum_{i=1}^{k}(x_i)}{k}$ as the mean. Finally, $TrueE$ is the population mean of the performance scores, denoted as $\mu$.

\vspace{3mm}
\noindent \textbf{Statement 1: $EstTrueE$ is consistent.} From \Cref{eq:eq5}, by rewriting $EstTrueE$, we have $EstTrueE = \frac{1}{n} \cdot \sum_{i=1}^{k}(x_i) \cdot \frac{n}{k} = \Bar{x}$, which is an unbiased estimator of the average performance $TrueE$, i.e., $Bias(\Bar{x}) = 0$ or  $\lim_{k \to \infty} Bias(EstTrueE) = 0$ (1). Now, let's denote the variance of the performance scores as $\sigma^2$, then the variance of $EstTrueE$ is $Var(EstTrueE) = Var(\Bar{x}) = \frac{\sigma^2}{n}$ and $\lim_{k \to \infty} Var(EstTrueE) = 0$ (2). From (1) and (2), by the  Sufficient Condition for Consistency \cite{amemiya1985advanced}, we conclude that $EstTrueE$ is a consistent estimator.

\vspace{3mm}
\noindent \textbf{Statement 2: $FI_C$ value.} Let's denote $s^2 = \frac{1}{k-1}\sum_{i=1}^{k}(x_i-\Bar{x})^2$ as the sample variance of the performance scores $x_i$s. It is well-known that $\frac{\sqrt{k}(\Bar{x}-\mu)}{s} \sim t_{k-1}$. For estimating the population mean $\mu$ with finite population size $n$ and the type I error $\alpha$, we have the margin of error $\epsilon$:

\begin{equation} \label{eq:eq9}
\epsilon \geq t_{\alpha/2,k-1} \cdot \sqrt{\frac{n-k}{n} \cdot \frac{s^2}{k}}
\end{equation}

where $\frac{n-k}{n}$ is the finite population correction factor. \Cref{eq:eq9} is equivalent to:

\begin{equation}
k \geq \frac{n-k}{n} \cdot \left( \frac{ t_{\alpha/2,k-1} \cdot s}{\epsilon} \right)^2
\end{equation}

which yields

\begin{equation}
k \geq \frac{1}{\frac{1}{n} + \left( \frac{\epsilon}{ t_{\alpha/2,k-1} \cdot s} \right)^2}.
\end{equation}

then

\begin{equation}
FI_C = \frac{k}{n} \geq \frac{1}{1 + n \cdot \left( \frac{\epsilon}{ t_{\alpha/2,k-1} \cdot s} \right)^2}.
\end{equation}

\vspace{3mm}
\noindent \textbf{Statement 3: When $FI_C$ approaches $1$, $EstTrueE$ approaches $TrueE$.} Since $EstTrueE$ by its definition in \Cref{eq:eq5} is continuous with respect to $FI_C$ (\Cref{eq:eq5}), $S_C$ (\Cref{eq:eq3}) and $F_C$ (\Cref{eq:eq3}), therefore, we have the equality: \[
\lim_{{FI_C} \to 100} {(EstTrueE)} = EstTrueE(FI_C=100)= {TrueE}.
\] 
\end{proof}

\subsection{Python Codes for Computing Reliability}

\begin{lstlisting}[language=Python, caption=Python codes for computing the reliability of $EstTrueE$ with margin of errors 5\% performance with a significance level 5\%.]
import numpy as np
from scipy.stats import t
import math 

def compute_sample_variance(data):
    n = len(data)
    mean = np.mean(data)
    squared_deviations = [(x - mean) ** 2 for x in data]
    sample_variance = sum(squared_deviations) / (n - 1)
    return sample_variance

def is_estimator_reliable(num_FI, list_eval_scores, num_samples=200):
    ####### t-statistics #######
    alpha = 0.05  # 5% significance level
    df = num_FI - 1  # degrees of freedom
    alpha_two_tailed = alpha / 2
    t_statistic = t.ppf(1 - alpha_two_tailed, df)

    #######Compute MOE_FI #######
    epsilon = 0.05  # 5% margin of error
    s = math.sqrt(compute_sample_variance(list_eval_scores))
    return num_FI/num_samples > 1/(1 + num_samples * (epsilon/(t_statistic * s))**2)
\end{lstlisting}

\newpage
\section{Detailed Discussions}
We give the numerical results and discussions for all figures and points made in the main paper.

\subsection{Multiple-choice Question (MCQ) Discussions} \label{appdx:mcq-discussions}

We evaluate Gemma, Mistral, and ChatGPT on the MMLU and BBH datasets using two prompting techniques, Zero-shot (ZS) and Zero-shot Chain-of-Thought (ZS-CoT) (\Cref{sec:mcq}). The prompts are specified in \Cref{appdx:mcq-prompt-details}. 
We report the $FI_C, SysE, EstTrueE$ scores. The results are presented in \Cref{tab:mcq-results}. Additionally, \Cref{MCQ:values}, \Cref{tab:mcq-fi-among-formats}, and \Cref{tab:mcq-cot-vs-noncot} are the distillation results of \Cref{tab:mcq-results}:

\begin{enumerate}
    \item \Cref{MCQ:values}. For each model, we average its $EstTrueE$ performance overall benchmarks and prompting techniques. For each task, we average the $EstTrueE$ scores overall models and prompting techniques. The results of this table are plotted in  \Cref{fig:mcq_main} and discussed in \Cref{sec:mcq-findings}.
    \item \Cref{tab:mcq-fi-among-formats}. The purpose of this table is to compare the FI scores across formats. We average all the FI scores across models and tasks. 
    \item \Cref{tab:mcq-cot-vs-noncot}. The purpose of this table is to see whether CoT \cite{wei2022chain} mitigates format bias. We average all the $EstTrueE$ scores over all models and benchmarks for each ZS and ZS-CoT prompting method. 
\end{enumerate}


\begin{table*}[!htp]
\centering
\footnotesize
\scalebox{1}{
\begin{tabular}{lcc}
\toprule 
\textbf{MCQ type} & \textbf{Char.} & \textbf{Text.} \\
\midrule
& \textbf{MMLU} & \\
Gemma-7B-it (EstTrue-Acc) & {0.53} / {27.25} &  \textcolor{red}{8.10} / {18.63} \\
Gemma-7B-it (Systematic-Acc) & 0.12 / 10.32 & 0.17 / 4.86 \\
Gemma-7B-it (FI) & 22.47 / 37.87 &  2.10 / 26.09 \\
\midrule
Mistral-7B-it-v0.2 (EstTrue-Acc) & {46.14} / {49.31} & \textcolor{red}{8.37} / \textcolor{red}{8.52} \\
Mistral-7B-it-v0.2 (Systematic-Acc) & 41.59 / 45.94 & 0.17 / 0.19 \\
Mistral-7B-it-v0.2 (FI) & 90.12 / 93.16 & 2.03 / 2.23 \\
\midrule
Llama-3.1-8B-it (EstTrue-Acc) & 20.04 / 35.03 &  \textcolor{red}{0.00} / 47.66 \\
Llama-3.1-8B-it (Systematic-Acc) & 1.79 / 27.16 & 0.00 / 29.05 \\
Llama-3.1-8B-it (FI) & 8.93 / 77.54 &  0.00 / 60.95 \\
\midrule
ChatGPT (EstTrue-Acc) & {68.55} / {45.53} & {54.85} / {59.67} \\
ChatGPT (Systematic-Acc) & 66.20 / 42.22 & 12.71 / 26.31 \\
ChatGPT (FI) & 96.56 / 92.73 & 23.17 / 44.09 \\
\midrule
\midrule
& \textbf{BBH} & \\
Gemma-7B-it (EstTrue-Acc) & \textcolor{red}{42.11} / {23.05} &  \textcolor{red}{0.00} / {15.11} \\
Gemma-7B-it (Systematic-Acc) & 0.40 / 13.00 & 0.00 / 6.80 \\ 
Gemma-7B-it (FI) & 0.95 / 56.40 &  0.00 / 45.00 \\
\midrule
Mistral-7B-it-v0.2 (EstTrue-Acc) & {76.81} / {62.50} & \textcolor{red}{0.00} / \textcolor{red}{0.00} \\
Mistral-7B-it-v0.2 (Systematic-Acc) & 21.20 / 22.00 & 0.00 / 0.00 \\
Mistral-7B-it-v0.2 (FI) & 27.60 / 35.20 & 0.00 / 1.60 \\
\midrule
Llama-3.1-8B-it (EstTrue-Acc) & \textcolor{red}{0.00} / 63.04 &  \textcolor{red}{0.00} / 48.40 \\
Llama-3.1-8B-it (Systematic-Acc) & 0.00 / 34.80 & 0.00 / 36.40 \\
Llama-3.1-8B-it (FI) & 0.00 / 55.20 &  0.00 / 75.20 \\
\midrule
ChatGPT (EstTrue-Acc) & {73.03} / {57.14} & {53.63} / \textcolor{red}{0.00} \\
ChatGPT (Systematic-Acc) & 26.00 / 16.0 & 53.20 / 0.00  \\
ChatGPT (FI) & 35.60 / 28.00 & 99.20 / 0.00 \\
\bottomrule
\end{tabular}}
\caption{\small{MCQ results. \textcolor{red}{Red} results are unreliable results measured by \Cref{theorem:theorem1} i.e., inequality \Cref{eq:eq6} does not hold.}}
\label{tab:mcq-results}
\end{table*}

\begin{table*}[!htp]
\centering
\scalebox{1}{
\begin{tabular}{lcc|c}
\toprule
 & \textbf{Char.} & \textbf{Text.} & \textbf{$BiasF_o$ (Var)} \\
\midrule
\multicolumn{3}{c}{\textbf{Models}} \\
\midrule
 Gemma & 23.24 & 10.46 & \textbf{40.83} \\
 Mistral & 58.69 &  4.22 & 741.74 \\
 Llama & 5.01 & 48.53 & 473.56 \\
 ChatGPT & 61.07 & 42.04 & 90.53 \\
\midrule
\multicolumn{3}{c}{\textbf{Tasks}} \\
\midrule
 MMLU & 32.17 & 30.10 & \textbf{1.06} \\
 BBH & 41.83 & 22.52 & 93.18 \\
\bottomrule
\end{tabular}}
\caption{\small{Average. estimated true accuracy results of MCQ
benchmarks across models and datasets (MMLU, BBH) with two MCQ formats (Character and Textual Value). This table is the distillation result of averaging all the estimated true performance scores across models (\textbf{Models}) and tasks (\textbf{Tasks}) in \Cref{MCQ:values}.
}}
\label{MCQ:values}
\end{table*}

\paragraph{Format instruction-following bias.} The FI scores across formats are illustrated in \Cref{tab:mcq-fi-among-formats}. There is a notable difference between the scores for character-based and textual value-based formats. Among the models, ChatGPT follows the instructions best with FI score $52.42\%$. Below we present two examples of Gemma and Mistral failing to follow the format instructions:

\begin{table*}[!htp]
\centering
\scalebox{1}{
\begin{tabular}{c|cc|cc}
\toprule
\textbf{Models} & \textbf{Char.} & \textbf{Text.} & \textbf{Avg.} & \textbf{$BiasF_o$ (Var)}\\
\midrule
Gemma & 29.42 & 18.29 & 23.86 & \textbf{30.96} \\
Mistral & 61.52 & 1.46 & 31.49 & 901.80 \\
Llama & 2.23 &  67.22 & 34.73 & 1055.92 \\
ChatGPT & \textbf{63.22} & \textbf{41.61} & \textbf{52.41} & 116.74 \\
\midrule
\textbf{Avg.} & 39.10 & 25.85 & 32.48 & 87.72 \\
\bottomrule
\end{tabular}}
\caption{\small{MCQ format-instruction following scores across formats. This table is the distillation result of averaging all the FI scores across tasks and prompting techniques in \Cref{MCQ:values}.
}}
\label{tab:mcq-fi-among-formats}
\end{table*}

\begin{tcolorbox}[colback=white]
\texttt{Answer the following question by outputting the textual value of the choice without its character identifier. \textcolor{blue}{Wrap your final answer by <ANSWER> and </ANSWER>.}}\\
\\
\texttt{Question: Is the following sentence plausible? ``John Carlson scored in the third period.''}\\
\texttt{Answer choices: A. yes | B. no}\\
\\
\texttt{\textbf{Gemma:}} \texttt{\textcolor{red}{<ANSWER>A</ANSWER>}}
\end{tcolorbox}

\begin{tcolorbox}[colback=white]
\texttt{Answer the following question by outputting the textual value of the choice without its character identifier. \textcolor{blue}{Wrap your final answer by <ANSWER> and </ANSWER>.}}\\
\\
\texttt{Question: Is the following sentence plausible? ``Marcelo got on the end of a through ball.''} \\
\texttt{Answer choices: A. yes | B. no}\\
\\
\texttt{\textbf{Mistral:}} \texttt{Based on the given sentence and the provided answer choices, the answer is: \textcolor{red}{A. yes}. The sentence describes Marcelo receiving a through ball and getting on the end of it, which is a common football (soccer) term for receiving a long, accurate pass and finishing a move with a shot or a goal.}
\end{tcolorbox}


\paragraph{ZS versus ZS-CoT.}
We investigate the impact of Chain-of-Thought (CoT) \cite{wei2022chain} on mitigating format bias. The average of estimated True Accuracy scores across models and tasks for the ZS prompting and ZS-CoT prompting are shown in \Cref{tab:mcq-cot-vs-noncot}. ZS prompting achieves a higher score in the character category (40.90\%) compared to ZS-CoT prompting (33.10\%). Similarly, for textual value format, ZS prompting scores higher than ZS-CoT prompting. However, the \textbf{$BiasF_o$} is lower for the ZS-CoT model ($17.42\%^2$) compared to the ZS model ($42.42\%^2$), indicating that CoT slightly decreases the format bias.

\begin{table*}[!htp]
\centering
\scalebox{1}{
\begin{tabular}{lcc|c}
\toprule
 & \textbf{Char.} & \textbf{Text.} & \textbf{$BiasF_o$}\\
\midrule
Zero-shot & 40.90 & 27.88 & 42.42 \\
Zero-shot Chain-of-Thought & 33.10 & 24.75 & \textbf{17.42} \\
\bottomrule
\end{tabular}}
\caption{\small{MCQ CoT versus non-CoT. This table is the distillation result of averaging all the Zero-shot and Zero-shot Chain-of-Thought scores across models and tasks in \Cref{MCQ:values}.}}
\label{tab:mcq-cot-vs-noncot}
\end{table*}


\paragraph{Reliability of the results.}
From \Cref{tab:mcq-results}, we see that $21/32$ of the estimated $EstTrue$ results are reliable. The reliability of results in the MCQ output format varies across different models. Gemma-7B-it and Mistral-7B-it show significant unreliability in textual value format, evidenced by numerous red-marked scores due to models not following the format instructions to output correct formats. In contrast, ChatGPT's results are significantly more reliable in the MMLU and BBH benchmarks ($7/8$), with only one unreliable result in the BBH textual format output. 


\subsection{Wrapping Discussions} \label{appdx:wrapping-discussions}

We examine Gemma, Mistral, and ChatGPT on the MCQ datasets (MMLU,BBH) and generation datasets (GSM8K, HotpotQA, FairytaleQA) utilizing two prompting techniques, Zero-shot (ZS) and Zero-shot Chain-of-Thought (ZS-CoT) (\Cref{sec:wrapping}). The prompts are also provided in \Cref{appdx:wrapping-prompt-details}. We measure the $FI_C, Sys E, EstTrue E$. The results are shown in \Cref{tab:wrapping-results}. Furthermore, \Cref{Wrapping:values}, \Cref{tab: wrapping FI-summary} and \Cref{Wrapping:Zs vs Zs-CoT} are the distillation outcome of \Cref{tab:wrapping-results}:


\begin{enumerate}
    \item \Cref{Wrapping:values}. For each model, we average its $EstTrueE$ performance overall benchmarks and prompting techniques. For each task, we average the $EstTrueE$ scores overall models and prompting techniques. This table is plotted in \Cref{fig:wrapping-bias} and discussed in \Cref{ssec:wrapping-findings}.
    \item \Cref{tab: wrapping FI-summary}. The purpose of this table is to compare the FI scores across formats. We average all the FI scores across models and tasks. 
    \item \Cref{Wrapping:Zs vs Zs-CoT}. The purpose of this table is to see whether CoT \cite{wei2022chain} mitigates format bias. We average all the $EstTrueE$ scores over all models and benchmarks for each ZS and ZS-CoT prompting method. 
\end{enumerate}


\begin{table*}[!htp]
\centering
\footnotesize
\scalebox{.75}{
\begin{tabular}{lccccccc}
\toprule 
\textbf{Wrapping type} & \textbf{Special character} & \textbf{Bolding} &  \textbf{Italicizing} & \textbf{Brackets} & \textbf{Parentheses} & \textbf{Placeholder} & \textbf{Quoting} \\
\midrule
& & & \textbf{MMLU} & & \\
Gemma-7B-it (EstTrue-Acc) & 35.59 / 20.28 & 41.28 / 44.27 & 49.85 / 74.18 & 36.36 / 32.95 & 36.68 / 20.12 & 46.45 / 25.77 & \textcolor{red}{60.41} / \textcolor{red}{74.06} \\
Gemma-7B-it (Systematic-Acc) & 27.82 / 20.28 & 21.66 / 17.73 & 26.64 / 27.89 & 28.55 / 27.28 & 10.53 / 12.96 & 29.80 / 21.96 & 2.64 / 2.37\\
Gemma-7B-it (FI) & 78.16 / 100.00 & 52.47 / 39.60 & 53.44 / 37.60 & 78.52 / 82.80 & 28.71 / 64.40 & 64.15 / 85.20 & 4.37 / 3.20 \\
\midrule
Mistral-7B-it (EstTrue-Acc) & 53.63 / 58.34 & \textcolor{red}{48.43} / \textcolor{red}{63.09}  & \textcolor{red}{51.84} / \textcolor{red}{61.66} &  67.36 / 61.58 & 64.99 / 62.71 & \textcolor{red}{75.35} / \textcolor{red}{6.03} &  \textcolor{red}{100.00} / \textcolor{red}{8.33}  \\
Mistral-7B-it (Systematic-Acc) & 13.42 / 20.04 & 1.08 / 9.40 & 4.80 / 10.15 & 20.08 / 17.28 & 11.10 / 13.42 & 1.07 / 0.14 & 0.03 / 0.01  \\
Mistral-7B-it (FI) & 23.81 / 34.35 & 2.23 / 14.90 & 9.26 / 16.46 & 29.81 / 28.06 & 17.08 / 21.40 & 1.42 / 2.32 & 0.03 / 0.12  \\
\midrule
Llama-3.1-8B-it (EstTrue-F1) & 52.06 / 55.95 & 25.08 / 54.42 &  48.54 / 89.11 & \textcolor{red}{92.36} / 19.63 & 39.28 / 32.00 & 65.18 / 68.12 & 34.67 / 42.83  \\
Llama-3.1-8B-it (Systematic-F1) & 38.05 / 37.92 & 12.93 / 28.82  &  20.48 / 32.57 & 11.61 / 8.02 & 31.80 / 30.39 & 56.53 / 27.91 & 26.89 / 16.91  \\
Llama-3.1-8B-it (FI) & 73.09 / 67.78 & 51.56 / 52.95 &  42.19 / 36.55 & 12.57 / 40.85 & 80.95 / 95.00 & 86.73 / 40.97 & 60.26 / 39.48  \\
\midrule
ChatGPT (EstTrue-Acc) &  54.64/ 71.28  & 67.40 / 75.86 & 44.76 / 64.79 &  59.80 / 71.42 & 57.82 / 71.11 & 66.24 / 72.81 &  68.29 / 70.68  \\
\textcolor{blue}{ChatGPT (Systematic-Acc)} & \textcolor{blue}{48.54} / 63.64 & \textcolor{blue}{66.59} / 48.59 & \textcolor{blue}{38.24} / 36.77 & \textcolor{blue}{31.65} / 60.86 & \textcolor{blue}{28.54} / 60.57 & \textcolor{blue}{63.88} / 50.09 & \textcolor{blue}{26.72} / 30.26  \\
ChatGPT (FI) & 88.84 / 89.28  & 98.80 / 64.05 & 85.43 / 56.75 & 52.93 / 85.21 & 49.36 / 85.18 & 96.44 / 68.80 & 39.13 / 42.81  \\
\midrule
\midrule
& & & \textbf{BBH} & & \\
Gemma-7B-it (EstTrue-Acc) & 25.00 / 16.00 & 49.09 / 38.38 & 52.94 / 24.47 & 63.04 / 47.34 & 36.73 / 26.09 & 7.07 / 3.76 & \textcolor{red}{60.00} / \textcolor{red}{20.00} \\
Gemma-7B-it (Systematic-Acc) & 24.00 / 16.00 & 21.60 / 15.20 & 10.80 / 9.20 & 23.20 / 19.60 & 14.40 / 16.80 & 5.20 / 3.20 & 2.40 / 0.40 \\
Gemma-7B-it (FI) & 96.00 / 100.00 & 44.00 / 39.60 & 20.40 / 37.60 & 36.80 / 41.40 & 39.20 / 64.40 & 73.60 / 85.20 & 4.00 / 2.00 \\
\midrule
Mistral-7B-it (EstTrue-Acc) & 52.40 / 64.00 & 10.40 / 11.60 & 36.80 / 21.20 & 16.00 / 8.40 & 6.4 / 12.00 & 32.80 / 72.80 & \textcolor{red}{0.00} / \textcolor{red}{0.00}  \\
Mistral-7B-it (Systematic-Acc) & 49.04 / 58.11 & 1.37 / 1.85 & 34.88 / 14.24 & 6.84 / 1.61 & 1.51 / 3.98 & 13.38 / 71.05 & 0.00 / 0.00  \\
Mistral-7B-it (FI) & 93.60 / 90.80 & 13.20 / 16.00 & 94.80 / 67.20 & 42.80 / 19.20 & 23.60 / 33.20 & 40.80 / 97.60 & 0.00 / 0.00  \\
\midrule
Llama-3.1-8B-it (EstTrue-F1) & 57.00 / 51.41 & 14.03 / 45.14 &  9.90 / 34.74 & 42.86 / 27.97 & 50.40 / 29.32 & 57.14 / 66.32 & 49.64 / 21.00  \\
Llama-3.1-8B-it (Systematic-F1) & 48.80 / 36.40 & 7.20 / 26.00 & 4.00 / 13.20 & 12.00 / 8.00 & 50.40 / 29.20 & 52.80 / 25.20 & 27.20 / 18.40  \\
Llama-3.1-8B-it (FI) & 85.60 / 70.80 & 53.60 / 57.60 & 40.80 / 38.00   & 28.00 / 27.60 & 100.00 / 99.60 & 92.40 / 38.00 & 54.80 / 37.60  \\
\midrule
ChatGPT (EstTrue-Acc) & 64.00 / 47.20 & 74.80 / 36.80 & 9.20 / 14.40 & 53.60 / 51.60 & 63.60 / 13.60 & 54.00 / 14.80 & 14.00 / \textcolor{red}{18.00}  \\
ChatGPT (Systematic-Acc) & 64.00 / 16.80 & 74.80 / 30.62 & 9.20 / 10.02 & 51.67 / 38.60 & 57.24 / 3.75 & 54.00 / 14.80 & 3.19 / 0.58  \\
ChatGPT (FI) & 100.00 / 35.60 & 100.00 / 83.20 & 100.00 / 69.60 & 96.40 / 74.80 & 90.00 / 27.60 & 100.00 / 100.00 & 22.80 / 3.20 \\
\midrule
\midrule
& & & \textbf{GSM8K} & & \\
Gemma-7B-it (EstTrue-$F1$) & 3.65 / 5.00 & 0.99 / 3.13 & 5.20 / 1.46 & 7.45 / 0.42 & \textcolor{red}{0.00} / \textcolor{red}{0.00} & 9.13 / 9.92 & \textcolor{red}{0.0} / \textcolor{red}{0.0} \\
Gemma-7B-it (Systematic-$F1$) & 2.54 / 2.45 & 0.50 / 2.00 & 4.26 / 1.19 & 3.50 / 0.17 & 0.00 / 0.00 & 4.52 / 4.71 & 0.0 / 0.0 \\
Gemma-7B-it (FI) & 69.50 / 49.00 & 50.50 / 64.00 & 82.00 / 81.50 & 47.00 / 40.05 & 2.50 / 0.50 & 49.50 / 47.50 & 0.0 / 0.0 \\
\midrule
Mistral-7B-it (EstTrue-F1) & 4.03 / 25.74 & 9.03 / \textcolor{red}{31.61} & 2.87 / 30.76 & 2.57 / 46.98 & 1.29 / 39.44 & 3.28 / 39.37 & \textcolor{red}{0.00} / \textcolor{red}{73.52}  \\
Mistral-7B-it (Systematic-F1) & 3.43 / 23.43 & 1.40 / 4.11 & 1.42 / 20.76 & 1.67 / 38.76 & 0.60 / 24.26 & 3.28 / 38.78 & 0.00 / 6.25  \\
Mistral-7B-it (FI) & 85.00 / 91.00 & 15.50 / 13.00 & 49.50 / 67.50 & 65.00 / 82.50 & 46.50 / 61.50 & 100.00 / 98.50 & 5.00 / 8.50  \\
\midrule
Llama-3.1-8B-it (EstTrue-F1) & 48.22 / 50.76 & 12.90 / 26.24 &  8.13 / 13.17 & 7.49 / 49.59 & 6.67 / 70.50 & 7.50 / 58.29 & \textcolor{red}{0.00} / \textcolor{red}{0.00}  \\
Llama-3.1-8B-it (Systematic-F1) & 47.50 / 50.00 & 10.00 / 18.50 &  5.00 / 11.00 & 7.00 / 30.50 & 5.50 / 49.00 & 7.50 / 58.00 & 0.00 / 0.00  \\
Llama-3.1-8B-it (FI) & 98.50 / 98.50 & 77.50 / 70.50 &  61.50 / 83.50 & 93.50 / 61.50 & 82.50 / 69.50 & 100.00 / 99.50 & 0.00 / 0.00  \\
\midrule
ChatGPT (EstTrue-F1) & 19.54 / 43.98 & 22.95 / 24.36 &  21.22 / 30.57 & 21.27 / 69.00 & 22.02 / 63.83 & 23.03 / 60.25 & 16.43 / 24.01  \\
ChatGPT (Systematic-F1) & 19.44 / 43.98 & 22.84 / 23.39 & 21.12 / 24.15 & 20.74 / 67.62 & 21.25 / 62.24 & 23.03 / 59.05 & 9.78 / 14.65 \\
ChatGPT (FI) & 99.50 / 100.00 & 99.50 / 96.00 & 99.50 / 79.00 & 97.50 / 98.50 & 96.50 / 97.50 & 100.00 / 98.00 & 59.50 / 61.00  \\
\midrule
\midrule
& & & \textbf{HotpotQA} & & \\
Gemma-7B-it (EstTrue-$F1$) & 14.12 / 9.88 & 21.43 / 32.11 & 19.83 / 27.06 & 23.63 / 30.44 & \textcolor{red}{0.00} / \textcolor{red}{0.00} & \textcolor{red}{43.70} / \textcolor{red}{53.62} & \textcolor{red}{2.33} / \textcolor{red}{6.60} \\
Gemma-7B-it (Systematic-$F1$) & 4.59 / 5.53 & 9.00 / 12.20 & 7.93 / 8.93 & 3.90 / 14.00 & 0.00 / 0.00 & 5.90 / 9.92 & 0.03 / 0.03 \\
Gemma-7B-it (FI) & 32.50 / 56.00 & 42.00 / 38.00 & 40.00 / 33.00 & 16.50 / 46.00 & 3.50 / 2.50 & 13.50 / 18.50 & 1.50 / 0.50 \\
\midrule
Mistral-7B-it (EstTrue-F1) & 12.86 / 11.43 & 25.84 / \textcolor{red}{29.21} & 20.93 / 14.56 & 16.93 / 13.20 & 15.39 / 13.21 & 20.41 / 21.58 & \textcolor{red}{0.00} / \textcolor{red}{25.00}  \\
Mistral-7B-it (Systematic-F1) & 7.27 / 3.83 & 8.27 / 3.36 & 6.91 / 4.95 & 16.51 / 10.76 & 14.55 / 10.24 & 19.70 / 14.75 & 0.00 / 0.05  \\
Mistral-7B-it (FI) & 56.50 / 33.50 & 32.00 / 11.50 & 33.00 / 34.00 & 97.50 / 81.50 & 94.50 / 77.50 & 96.50 / 91.50 & 0.00 / 0.20  \\
\midrule
Llama-3.1-8B-it (EstTrue-F1) & 20.63 / 21.65 & 21.93 / 14.11 &  5.34 / 5.92 & 19.17 / 18.08 & 19.79 / 17.59 & 20.00 / 21.39 & \textcolor{red}{0.00} / \textcolor{red}{0.00}  \\
Llama-3.1-8B-it (Systematic-F1) & 13.00 / 10.50 & 21.50 / 11.50 &  5.00 / 4.00 & 18.50 / 8.50 & 19.50 / 9.50 & 20.00 / 20.00 & 0.00 / 0.00  \\
Llama-3.1-8B-it (FI) & 63.00 / 48.50 & 98.00 / 81.50 &  93.50 / 67.50 & 96.50 / 47.00 & 98.50 / 54.00 & 100.00 / 93.50 & 0.00 / 0.00  \\
\midrule
ChatGPT (EstTrue-F1) & 29.86 / 27.52 & 41.00 / 33.14 & 35.39 / 28.96 & 23.94 / 35.48 & 29.30 / 34.83 & 38.72 / 28.69 & 41.52 / 16.97 \\
ChatGPT (Systematic-F1) & 25.24 / 27.11 & 40.59 / 30.82 & 33.45 / 26.64 & 17.00 / 33.36 & 23.46 / 33.44 & 38.72 / 27.69 & 11.73 / 7.13  \\
ChatGPT (FI) & 84.50 / 98.50 & 99.00 / 93.00 & 94.50 / 92.00 & 71.50 / 94.00 & 80.05 / 96.00 & 100.00 / 96.50 & 28.50 / 42.00  \\
\midrule
\midrule
& & & \textbf{FairytaleQA} & & \\
Gemma-7B-it (EstTrue-$F1$) & 17.42 / 29.72 & 8.91 / 0.97 & 8.12 / 14.50 & \textcolor{red}{22.13} / \textcolor{red}{18.62} & \textcolor{red}{0.00} / \textcolor{red}{0.00} & \textcolor{red}{20.64} / 22.05 & \textcolor{red}{0.00} / \textcolor{red}{0.00} \\
Gemma-7B-it (Systematic-$F1$) & 6.62 / 11.74 & 4.68 / 0.64 & 4.75 / 9.79 & 1.77 / 1.21 & 0.00 / 0.00 & 2.58 / 4.08 & 0.00 / 0.00 \\
Gemma-7B-it (FI) & 38.00 / 39.50 & 52.50 / 66.00 & 58.50 / 67.50 & 8.00 / 6.50 & 0.00 / 0.00 & 12.50 / 18.50 & 0.00 / 0.00 \\
\midrule
Mistral-7B-it (EstTrue-F1) & 27.19 / 22.20 & \textcolor{red}{23.78} / \textcolor{red}{50.00} & 47.36 / 29.49 & 32.42 / 25.90 & 30.33 / 22.46 & 36.07 / 31.77 & \textcolor{red}{19.50} / \textcolor{red}{20.00}  \\
Mistral-7B-it (Systematic-F1) & 22.16 / 18.54 & 3.21 / 0.50 & 18.47 / 15.19 & 32.42 / 25.00 & 29.73 / 21.00 & 35.89 / 31.62 & 0.39 / 1.30  \\
Mistral-7B-it (FI) & 81.50 / 83.50 & 13.50 / 1.00 & 39.00 / 51.50 & 100.00 / 96.50 & 98.00 / 93.50 & 99.50 / 99.50 & 2.00 / 6.50  \\
\midrule
Llama-3.1-8B-it (EstTrue-F1) & 49.62 / 36.78 & 48.35 / 39.96 &  45.42 / 35.75 & 48.85 / 40.76 & 48.63 / 42.48 & 52.55 / 40.39 & \textcolor{red}{0.00} / \textcolor{red}{0.00}  \\
Llama-3.1-8B-it (Systematic-F1) & 48.13 / 36.41 & 48.35 / 36.76 &  45.42 / 29.85 & 48.85 / 20.79 & 48.39 / 26.55 & 52.29 / 39.18 & 0.00 / 0.00  \\
Llama-3.1-8B-it (FI) & 97.00 / 99.00 & 100.00 / 92.00 &  100.00 / 83.50 & 100.00 / 51.00 & 99.50 / 62.50 & 99.50 / 97.00 & 0.00 / 0.00  \\
\midrule
ChatGPT (EstTrue-F1) & {41.93} / {31.95} & 46.08 / 32.84 &  48.11 / {33.46} & {41.53} / {38.25} & {38.25} / {34.82} & {46.83} / {32.85} & {45.78} / {27.75}  \\
ChatGPT (Systematic-F1) & 38.58 / 31.47 & 46.08 / 31.86 & 48.11 / 31.96 & 41.33 / 38.06 & 45.91 / 34.30 & 46.83 / 32.85 & 27.24 / 14.71  \\
ChatGPT (FI) & 92.00 / 98.50 & 100.00 / 97.00 & 100.00 / 95.50 & 99.50 / 99.50 & 99.50 / 98.50 & 100.00 / 100.00 & 59.50 / 53.00  \\
\bottomrule
\end{tabular}
}
\caption{\small{Wrapping results. \textcolor{red}{Red} results are unreliable results measured by \Cref{theorem:theorem1} i.e., inequality \Cref{eq:eq6} does not hold.}}
\label{tab:wrapping-results}
\end{table*}

\begin{table*}[!htp]
\centering
\footnotesize
\scalebox{.8}{
\begin{tabular}{lccccccc|c}
\toprule
 & \textbf{Special Character} & \textbf{Bolding} & \textbf{Italicizing} & \textbf{Brackets} & \textbf{Parentheses} & \textbf{Placeholder} & \textbf{Quoting} & \textbf{$BiasF_o$ (Var)} \\
\midrule
\multicolumn{8}{c}{\textbf{Models}} \\
\midrule
Gemma & 17.69 & 22.23 & 28.24 & 23.35 & 9.86 & 25.88 & 22.25 & 31.31 \\
Mistral & 32.87 & 29.33 & 31.05 & 28.46 & 26.02 & 33.23 & 24.64 & 9.20 \\
LLama & 44.41 & 30.16 & 28.98 & 36.78 & 35.67 & 45.69 & 18.85 & 74.86 \\
ChatGPT & 40.15 & 44.91 & 32.65 & 45.24 & 42.72 & 43.82 & 32.57 & 26.03 \\
\midrule
Average & 33.78 & 31.65 & 30.23 & 33.46 & 28.57 & 37.15 & 24.58 & 14.14 \\
\midrule
\multicolumn{8}{c}{\textbf{Tasks}} \\
\midrule
MMLU & 50.56 &	52.48 & 60.59 &	55.18 &	48.09 & 53.24 & 58.65 & 16.58 \\
BBH & 47.13 & 34.96 & 24.68 & 38.98 & 29.77 & 38.59 & 26.32 & 54.58 \\
GSM8K & 25.12 & 16.40 & 14.17 & 25.55 & 25.47 & 26.35 & 14.25 & 28.51 \\
HotpotQA & 18.50 & 27.35 & 19.75 & 22.59 & 16.27 & 29.41 & 11.39 & 33.66 \\
FairytaleQA & 32.10 & 31.36 & 32.78 & 30.79 & 28.11 & 35.40 & 14.13 & 42.16 \\
\midrule
Average & 34.68 & 32.51 & 30.39 & 34.62	& 29.54	& 36.60	& 24.95 & 13.39 \\
\bottomrule
\end{tabular}}
\caption{\small{Avg. estimated true accuracy results of 
benchmarks across models and datasets with seven Wrapping formats (Special Character, Bolding, Italicizing, Brackets, Parentheses, Placeholder, Quoting). This table is the distillation result of averaging all the estimated true performance scores across models and benchmarks in \Cref{tab:wrapping-results}.}}
\label{Wrapping:values}
\end{table*}

\paragraph{Format instruction-following bias.}
The FI scores over formats are provided in \Cref{tab: wrapping FI-summary}. Overall, LLMs exhibit significant format-following bias across formats with a variance of FI scores of $345.85 \%^2$. Among the models, ChatGPT follows the instructions best with an average FI Score of $85.01\%$. The ``Special Character'' wrapping format has the highest FI score of $75.05\%$. Following it is the ``Placeholder'' wrapping format also shows a high FI score of $72.47\%$, suggesting it is another effective format for ensuring instruction adherence. In contrast, the ``Quoting'' wrapping format has the lowest FI score of $18.55\%$. This significant drop compared to other formats suggests that quoting is the least effective method for wrapping instructions, possibly causing confusion or misinterpretation by the models. Below we present two examples of Gemma and Mistral failing to follow the format instructions:


\begin{table*}[!htp]
\centering
\footnotesize
\scalebox{.85}{
\begin{tabular}{c|ccccccc|cc}
\toprule
 \textbf{Model} & \textbf{Special Character} & \textbf{Bolding} & \textbf{Italicizing} & \textbf{Brackets} & \textbf{Parentheses} & \textbf{Placeholder} & \textbf{Quoting} & \textbf{Avg.} & \textbf{$BiasF_o$ (Var)} \\
\midrule
Gemma & 62.27 & 46.07 & 49.51 & 37.06 & 15.33 & 45.58 & 1.36 & 36.74 &  384.31 \\ 
Mistral & 63.00 & 15.36 & 40.94 & 67.53 & 54.72 & 64.48 & 4.04 & 44.30	& 553.55 \\ 
LLama & 80.18 & 73.52 & 64.70 & 55.85 & 84.21 & 84.76 & 23.01 & 66.60 & 413.00 \\
ChatGPT & \textbf{94.77} & \textbf{93.49} & \textbf{88.35} & \textbf{88.93} & \textbf{88.69} & \textbf{95.06} & \textbf{45.79} & \textbf{85.01} & \textbf{263.71} \\ 
\midrule
\textbf{Avg.} & 75.05 & 57.11 & 59.60 & 60.87 & 60.74 & 72.47 & 18.55 & 57.76 & 345.85\\
\bottomrule
\end{tabular}}
\caption{\small{Avg. Following Instruction Score over all the wrapping formats. This table is the distillation result of averaging all the FI scores across models and benchmarks in \Cref{tab:wrapping-results}.
}}
\label{tab: wrapping FI-summary}
\end{table*}

\begin{tcolorbox}[colback=white]
\texttt{Answer the following question without any explanation. \textcolor{blue}{Wrap your final answer using triple quotation marks.}}\\
\\
\texttt{Question (HotpotQA): ``What was the MGM Grand Garden Arena in which Britney Spears recorded fourth video album originally known as?''} \\ 
\texttt{Context:...}\\
\\
\texttt{\textbf{Gemma:} \textcolor{red}{MGM Grand Garden Special Events}}.
\end{tcolorbox}

\begin{tcolorbox}[colback=white]
\texttt{Answer the following question without any explanation. \textcolor{blue}{Wrap your final answer using triple quotation marks.}}\\
\\
\texttt{Question (BBH): Is the following sentence plausible? ``David Silva took a throw-in.''}\\ 
\\
\texttt{\textbf{Mistral:} \textcolor{red}{A. ``yes''}}.
\end{tcolorbox}


\paragraph{ZS versus ZS-CoT.}  The average of estimated True Accuracy scores across models and tasks for the ZS and ZS-CoT prompting are shown in  \Cref{Wrapping:Zs vs Zs-CoT}. For the majority of the wrapping methods (``Special Character'', ``Bolding'', ``Italicizing'', and ``Brackets''), the ZS-CoT model generally shows higher or comparable performance to the ZS model. The ``Italicizing'' shows a significant improvement when using ZS-CoT, with a jump from 28.28\% (ZS) to 32.19\% (ZS-CoT). However, the \textbf{$BiasF_o$} metric shows a considerable difference between $11.39\%^2$ (ZS) and $16.22\%^2$ (ZS-CoT). This depicts that while applying CoT may improve accuracy in certain methods, it does not generally reduce format bias.

\begin{table*}[!htp]
\centering
\footnotesize
\scalebox{.85}{
\begin{tabular}{lccccccc|c}
\toprule
 & \textbf{Special Character} & \textbf{Bolding} & \textbf{Italicizing} & \textbf{Brackets} & \textbf{Parentheses} & \textbf{Placeholder} & \textbf{Quoting} & \textbf{$BiasF_o$}\\
\midrule
Zero-shot & 34.31 & 30.66 & 28.28 & 33.82 & 28.77 & 35.93 & 26.09 & \textbf{11.39} \\
Zero-shot Chain-of-Thought & 33.25 & 32.65 & 32.19 & 33.09 & 28.36 & 38.37 & 24.46 & 16.22 \\
\bottomrule
\end{tabular}}
\caption{\small{Avg.Estimated Accuracy of non CoT versus CoT for wrapping methods. This table is the distillation result of averaging all the Zero-shot and Zero-shot Chain-of-Thought scores across models and tasks in \Cref{tab:wrapping-results}.}}
\label{Wrapping:Zs vs Zs-CoT}
\end{table*}


\paragraph{Reliability of the results.}
Overall, $80\%$ of the $EstTrue$ results are reliable. Gemma-7B-it shows mixed reliability, with some red-marked scores indicating unreliable results, particularly in the ``Quoting'' format. This is because Gemma failed to follow the quoting instruction to quote the final answer. Mistral-7B-it exhibits similar variability, with some unreliable scores in ``Quoting'' and ``Placeholder'' formats. ChatGPT generally demonstrates mostly reliable results, with only 1 quoting result unreliable. 



\subsection{List Discussions} \label{appdx:list-discussions}

\begin{table*}[!htp]
\centering
\footnotesize
\scalebox{1}{
\begin{tabular}{lcccc}
\toprule 
\textbf{Listing type} & \textbf{Python} & \textbf{Bullet} & \textbf{Spe. Char.} & \textbf{Newline} \\
\midrule
& & \textbf{SciDocsRR} &\\
Gemma-7B-it (EstTrue-$mAP$) & \textcolor{red}{0.0} / 61.65 & \textcolor{red}{0.0} / \textcolor{red}{73.0} & \textcolor{red}{0.0} / \textcolor{red}{60.00} & \textcolor{red}{0.0} / 60.15 \\
Gemma-7B-it (Systematic-$mAP$) & 0.0 / 15.72 & 0.0 / 1.46 & 0.0 / 0.90 & 0.0 / 28.27 \\
Gemma-7B-it (FI) & 0.0 / 25.50 & 0.0 / 2.00 & 0.0 / 1.50 & 0.0 / 47.00 \\
\midrule
Mistral (EstTrue-$mAP$) & 50.21 / 52.61  & \textcolor{red}{0.00} / \textcolor{red}{0.00} & \textcolor{red}{0.00} / \textcolor{red}{0.00} & 78.08 / 58.36 \\
Mistral (Systematic-$mAP$) & 37.41 / 9.47 & 0.00 / 0.00 & 0.00 / 0.00 & 18.35 / 27.14 \\
Mistral (FI) & 74.50 / 18.00 & 0.00 / 0.00 & 0.00 / 0.00 & 23.50 / 46.50 \\
\midrule
Llama-3.1-8B-it  (EstTrue-$F1$) & 32.10 / \textcolor{red}{0.00} & \textcolor{red}{38.80} / \textcolor{red}{0.00} & \textcolor{red}{0.00} / \textcolor{red}{0.00} & 35.78 / \textcolor{red}{0.00} \\
Llama-3.1-8B-it  (Systematic-$F1$) & 6.26 / 0.00 & 3.88 / 0.00 & 0.00 / 0.00 & 35.78 / 0.00 \\
Llama-3.1-8B-it  (FI) & 19.50 / 0.00 & 10.00 / 0.00 & 0.00 / 0.00 & 100.00 / 0.00 \\
\midrule
ChatGPT (EstTrue-$mAP$) & 35.29 / 50.17 & 49.94 / 59.64 & 55.69 / 57.78 & 38.54 / 57.56 \\
ChatGPT (Systematic-$mAP$) & 33.17 / 28.60 & 49.19 / 25.05 & 55.69 / 37.85 & 35.46 / 35.41\\
ChatGPT (FI) & 94.00 / 57.00 & 98.50 / 42.00 & 100.00 / 65.50 & 92.00 / 61.50 \\
\midrule
\midrule
& & \textbf{SemEval2017} &\\
Gemma-7B-it (EstTrue-$F1$) & \textcolor{red}{4.00} / \textcolor{red}{8.86} & 7.10 / 7.20 & 4.80 / 13.50 & 7.21 / 3.25 \\
Gemma-7B-it (Systematic-$F1$) & 0.04 / 1.64 & 1.80 / 2.10 & 4.80 / 13.50 & 7.21 / 1.51 \\
Gemma-7B-it (FI) & 1.00 / 18.50 & 25.50 / 29.15 & 100.00 / 100.00 & 100.00 / 46.50 \\
\midrule
Mistral (EstTrue-$F1$) & 34.82 / 30.24 & 23.2 / 0.00 & \textcolor{red}{0.00} / 13.57 & 12.17 / 20.84 \\
Mistral (Systematic-$F1$) & 33.95 / 24.19 & 23.20 / 0.00 & 0.00 / 10.72 & 12.17 / 20.84 \\
Mistral (FI) & 97.50 / 80.00 & 100.00 / 100.00 & 0.00 / 79.00 & 100.00 / 100.00 \\
\midrule
Llama-3.1-8B-it  (EstTrue-$F1$) & 34.82 / \textcolor{red}{0.00} & 0.22 / 0.00 & \textcolor{red}{0.00} / 0.00 & 0.00 / 10.17 \\
Llama-3.1-8B-it  (Systematic-$F1$) & 33.95 / 0.00 & 0.22 / 0.00 & 0.00 / 0.00 & 0.00 / 10.17 \\
Llama-3.1-8B-it  (FI) & 97.50 / 0.00 & 100.00 / 100.00 & 0.00 / 75.50 & 100.00 / 100.00 \\
\midrule
ChatGPT (EstTrue-$F1$) & 42.25 / 15.33 & 8.87 / 16.46 & 32.19 / 16.33 & 37.16 / 22.87 \\
ChatGPT (Systematic-$F1$) & 39.51 / 6.04 & 8.87 / 16.13 & 31.07 / 15.51 & 37.16 / 22.75 \\
ChatGPT (FI) & 93.50 / 39.39 & 100.00 / 97.97 & 96.50 / 94.94 & 100.00 / 99.49 \\
\bottomrule
\end{tabular}
}
\caption{\small{List results. \textcolor{red}{Red} results are unreliable results measured by \Cref{theorem:theorem1} i.e., inequality \Cref{eq:eq6} does not hold.}}
\label{tab:list-results}
\end{table*}

We assess Gemma, Mistral, and ChatGPT with two prompting
techniques, Zero-shot (ZS) and Zero-shot Chain-of-Thought (ZS-CoT) (\Cref{sec:list}) on two benchmarks SciDocsRR and SemEval2017. Our prompts are provided in \ref{appdx:list-prompt-details}. We utilize $FI_C, SysE, TrueE$ as our evaluation metrics. The results are illustrated in \Cref{tab:list-results}. In addition, \Cref{tab:list-values}, \Cref{tab:list-fi-among-formats} and \Cref{tab:list-zs-cot} are the distillation results of \Cref{tab:list-results}:

\begin{enumerate}
    \item \Cref{tab:list-values}. For each model, we average its $EstTrueE$ performance overall benchmarks and prompting techniques. For each task, we average the $EstTrueE$ scores overall models and prompting techniques. This table is drawn in \Cref{fig:list-bias} and its discussions are conducted in \Cref{sec:list}. 
    \item \Cref{tab:list-fi-among-formats}. The purpose of this table is to compare the FI scores across formats. We average all the FI scores across models and tasks.
    \item \Cref{tab:list-zs-cot}. The purpose of this table is to see whether CoT \cite{wei2022chain} mitigates format bias. We average all the $EstTrueE$ scores over all models and benchmarks for each ZS and ZS-CoT prompting method. 
\end{enumerate}

\begin{table*}[!htp]
\centering
\footnotesize
\scalebox{.9}{
\begin{tabular}{lcccc|c}
\toprule
 & \textbf{Python} & \textbf{Bullet} & \textbf{Special Character} & \textbf{Newline} & \textbf{$BiasF_o$ (Var)} \\
\midrule
\multicolumn{5}{c}{\textbf{Models}} \\
\midrule
Gemma & 17.12 & 18.25 & 15.12 & 16.21 & \textbf{1.32} \\
Mistral & 41.98 & 5.80 & 3.39 & 42.37 & 353.80 \\
LLama & 16.73 & 9.76 & 0.00 & 11.49 & 36.64 \\
ChatGPT & 35.76 & 33.73 & 40.50 & 39.03 & 7.08 \\
\midrule
Average & 27.90 & 16.88 & 14.75 & 27.27 & 46.97 \\
\midrule
\multicolumn{5}{c}{\textbf{Tasks}} \\
\midrule
SemEval2017 & 20.54 & 6.09 & 7.82 & 13.49 & \textbf{31.86} \\
SciDocsRR & 35.25 & 27.67 & 21.68 & 41.06 & 54.12 \\
\midrule
Average & 27.90	& 16.88 & 14.75 & 27.27 & 46.97 \\
\bottomrule
\end{tabular}}
\caption{Avg estimated true F1 (SemEval2017) and MAP
(SciDocsRR) scores over benchmarks (SemEval2017, SciDocsRR) and models (Gemma-7B-it, Mistral, ChatGPT) across 4 widely used list formats (Python, Bullet, Special Character, Newline)}
\label{tab:list-values}
\end{table*}

\paragraph{Format instruction-following bias.}
From \cref{tab:list-fi-among-formats}, the ``Newline list'' format has the highest FI score of $68.29\%$, while the FI scores of other formats are around $40-50\%$. Among the models, ChatGPT follows the instructions best with FI score $83.27\%$. Specifically, ChatGPT excels at following instructions for all formats. In comparison, Mistral performs poorly with ``Special character list'', while Gemma struggles the most with ``Python'' and ``Bullet point list''. Explaining these phenomena, we observe that Gemma often misinterprets the task of converting textual samples into lists of numbers as a coding task, resulting in answers that do not match the desired format or content. We take an example in dataset \textbf{SciDocsRR} \Cref{appdx:dataset} with detailed prompt as in \Cref{appdx:list-prompt-details}
\begin{tcolorbox}[colback=white]
\texttt{Given a query, and a list of documents.}\\
Query: \texttt{Declarative Terrain Modeling for Military Training Games} \\ 
Documents: \texttt{[...]} \\ 
\\
\texttt{You are required to transform the list of documents into a binary list of 1 or 0 where 1 indicates the document brings useful information to the topic, and 0 indicates the document does not bring useful information to the topic. \textcolor{blue}{Generate your binary list as a Python list}} \\
\\
\texttt{\textbf{Gemma:} \textcolor{red}{You should not use any packages such as sklearn, gensim, and NLTK. You are required to write a short explanation for your binary list.}}
\end{tcolorbox}

\begin{tcolorbox}[colback=white]
\texttt{Given a query, and a list of documents.}\\
Query: \texttt{Declarative Terrain Modeling for Military Training Games} \\ 
Documents: \texttt{[...]} \\ 
\\
\texttt{You are required to transform the list of documents into a binary list of 1 or 0 where 1 indicates the document brings useful information to the topic, and 0 indicates the document does not bring useful information to the topic. \textcolor{blue}{Generate your binary list using bullet points}} \\
\\
\texttt{\textbf{Gemma:} \textcolor{red}{Your binary list must be in the following format:
[1, 0, 1, 0, 0, 0, 1, 1, 0, 1, 0, 1, 1, 1, 1, 0, 1, 1, 1, 0]}}.
\end{tcolorbox}
\begin{table*}[!htp]
\centering
\footnotesize
\scalebox{1}{
\begin{tabular}{c|cccc|cc}
\toprule
\textbf{Model} & \textbf{Python} & \textbf{Bullet} & \textbf{Special Character} & \textbf{Newline} & \textbf{Avg.} & \textbf{$BiasF_o$ (Var)} \\
\midrule
Gemma & 7.51 & 13.41 &	34.55 &	42.40 & 24.46 & 277.69 \\
Mistral & 67.50 & 50.00 & 19.75 & 67.50 & 51.19 & 507.31 \\
ChatGPT & \textbf{70.97} & \textbf{84.61} & \textbf{89.24} &	\textbf{88.25} & \textbf{83.27} & \textbf{71.13} \\
LLama & 29.25 & 52.50 & 18.88 & 75 & 43.91 & 470.51 \\
\midrule
\textbf{Avg.} & 43.81 & 50.13 &	40.60 &	68.29 & 50.71 & 153.04\\
\bottomrule
\end{tabular}}
\caption{\small{Avg Following Instruction scores over benchmarks (SemEval2017, SciDocsRR) and models (Gemma-7B-it, Mistral, ChatGPT) across 4 widely used list formats (Python, Bullet, Special Character, Newline). This table is the distillation result of averaging all the
FI scores across models and benchmarks in \Cref{tab:list-results}.}}
\label{tab:list-fi-among-formats}
\end{table*}

\paragraph{ZS versus ZS-CoT.}
The results, detailed in \cref{tab:list-zs-cot} indicate that prompting with ZS-CoT substantially enhances model performance across various formats. Moreover, ZS-CoT effectively reduces format bias, as evidenced by the $BiasF_o$ metric decreasing from about $54\%^2$ to $22\%^2$. From this, we conclude that CoT reduces format bias.

\begin{table*}[!htp]
\centering
\footnotesize
\scalebox{1}{
\begin{tabular}{l|cccc|c}
\toprule
& \textbf{Python} & \textbf{Bullet} & \textbf{Special Character} & \textbf{Newline} & \textbf{$BiasF_o$ (Var)} \\
\midrule
Zero-shot & 29.19 & 15.13 & 11.05 & 25.45 & 54.48 \\
Zero-shot Chain-of-Thought & 26.61 & 18.64 & 18.46 & 29.09 & \textbf{22.40} \\		
\bottomrule
\end{tabular}}
\caption{\small{Avg estimated true F1 (SemEval2017) and MAP
(SciDocsRR) scores of non-CoT versus CoT for list formats. This table is the distillation result of averaging all the scores across models and benchmarks in \Cref{tab:list-results}.}}
\label{tab:list-zs-cot}
\end{table*}

\paragraph{Reliability of the results.}
From \cref{tab:list-results}, $67\%$ of the $EstTrue$ results are reliable. However, some scores of Gemma-7B-it and LLama on these benchmarks are red-marked, indicating unreliable results of this model. In contrast, the ChatGPT's results are perfectly reliable.

\subsection{Mapping Discussions} \label{appdx:mapping-discussions}

\begin{table*}[!htp]
\centering
\footnotesize
\scalebox{1}{
\begin{tabular}{lcc}
\toprule 
\textbf{Mapping type} & \textbf{JSON} & \textbf{YAML} \\
\midrule
& \textbf{SciREX  Easy} & \\
Gemma-7B-it (EstTrue-F1) & 14.60 / 20.84 & 18.20 / \textcolor{red}{0.82} \\
Gemma-7B-it (Systematic) & 3.54 / 3.79 & 3.03 / 0.10 \\
Gemma-7B-it (FI) & 24.24 / 18.18 & 16.64 / 12.12 \\
\midrule
Mistral-7B-it (EstTrue-F1) & 28.83 / 32.82 & \textcolor{red}{0.00} / \textcolor{red}{0.00} \\
Mistral-7B-it (Systematic) & 11.36 / 32.33 & 0.00 / 0.00 \\
Mistral-7B-it (FI) & 39.39 / 98.48 & 0.00 / 3.03 \\
\midrule
Llama-3.1-8B-it (EstTrue-F1) & \textcolor{red}{25.85} / 22.86 & 19.07 / 28.57 \\
Llama-3.1-8B-it (Systematic) & 2.35 / 14.55 & 6.36 / 15.15 \\
Llama-3.1-8B-it (FI) & 9.09 / 63.63 & 33.34 / 53.03 \\
\midrule
ChatGPT (EstTrue-F1) & 35.99 / 22.40 & 23.63 / 26.60 \\
ChatGPT (Systematic) & 32.72 / 19.69 & 22.92 / 20.15 \\
ChatGPT (FI) & 90.90 / 87.87 & 96.96 / 75.75 \\
\midrule
\midrule
& \textbf{SciREX  Medium} & \\
Gemma-7B-it (EstTrue-F1) & 18.17 / 5.27 & 0.00 / \textcolor{red}{1.87} \\
Gemma-7B-it (Systematic) & 3.03 / 0.88 & 0.00 / 0.17 \\
Gemma-7B-it (FI) & 16.67 / 16.67 & 18.18 / 9.09 \\
\midrule
Mistral-7B-it (EstTrue-F1) & 26.48 / 23.81 & \textcolor{red}{18.97} / \textcolor{red}{20.83} \\
Mistral-7B-it (Systematic) & 21.27 / 23.81 & 1.15 / 0.25 \\
Mistral-7B-it (FI) & 80.30 / 100.00 & 6.06 / 1.20 \\
\midrule
Llama-3.1-8B-it (EstTrue-F1) & 40.80 / 35.01 & 31.64 / 27.10 \\
Llama-3.1-8B-it (Systematic) & 9.89 / 28.12 & 25.41 / 21.76 \\
Llama-3.1-8B-it (FI) & 24.24 / 80.30 & 80.30 / 80.30\\
\midrule
ChatGPT (EstTrue-F1) & 29.07 / 27.29 & 36.55 / 22.70 \\
ChatGPT (Systematic) & 28.19 / 26.47 & 21.60 / 22.70 \\
ChatGPT (FI) & 96.96 / 96.96 & 59.09 / 100.00 \\
\midrule
\midrule
& \textbf{SciREX  Hard} & \\
Gemma-7B-it (EstTrue-F1) & \textcolor{red}{34.40} / 29.18 & 1.65 / \textcolor{red}{0.87} \\
Gemma-7B-it (Systematic) & 4.17 / 10.61 & 0.25 / 0.04 \\
Gemma-7B-it (FI) & 12.12 / 36.36 & 15.15 / 4.55 \\
\midrule
Mistral-7B-it (EstTrue-F1) & 22.44 / 30.34 & \textcolor{red}{12.54} / \textcolor{red}{15.95} \\
Mistral-7B-it (Systematic) & 20.40 / 26.66 & 1.71 / 1.58 \\
Mistral-7B-it (FI) & 90.90 / 87.87 & 13.63 / 9.90 \\
\midrule
Llama-3.1-8B-it (EstTrue-F1) & 39.33 / 39.66 & 27.39 / 26.05 \\
Llama-3.1-8B-it (Systematic) & 5.36 / 33.05 & 25.32 / 24.08 \\
Llama-3.1-8B-it (FI) & 13.63 / 92.42 & 83.33 / 92.42 \\
\midrule
ChatGPT (EstTrue-F1) & 20.25 / 22.57 & 11.76 / 12.07 \\
ChatGPT (Systematic) & 19.64 / 22.23 & 11.59 / 10.43 \\
ChatGPT (FI) & 96.96 / 98.48 & 98.48 / 86.36 \\
\bottomrule
\end{tabular}}
\caption{\small{Mapping results. \textcolor{red}{Red} results are unreliable results measured by \Cref{theorem:theorem1} i.e., inequality \Cref{eq:eq6} does not hold.}}
\label{tab:mapping-results}
\end{table*}

\begin{table*}[!htp]
\centering
\footnotesize
\scalebox{1}{
\begin{tabular}{lcc|cc}
\toprule
 & \textbf{JSON} & \textbf{YAML} & \textbf{Average} & \textbf{$BiasF_o$ (Var)}\\
\midrule
\multicolumn{3}{c}{\textbf{Models}} \\
\midrule
Gemma & 20.42 & 3.91 & 12.17 & 68.14 \\
Mistral & 27.46 &  11.39 & 19.43 & 64.56 \\
Llama & 33.92 & 26.64 & 30.28 & 13.25\\
ChatGPT & 26.27 & 22.22 & 24.25 & \textbf{4.10}\\
\midrule
\multicolumn{3}{c}{\textbf{Tasks}} \\
\midrule
Easy & 25.53 & 14.61 & 20.07 & 29.79\\
Medium & 25.74 & 19.96 & 22.85 & \textbf{8.36}\\
Hard & 29.77 & 13.54 & 21.66 & 65.87\\
\bottomrule
\end{tabular}}
\caption{\small{Avg estimated true F1 scores over benchmarks
(SciREX Easy, SciREX Medium and SciREX Hard) and models (Gemma-7B-it, Mistral, ChatGPT) across 2 widely used mapping formats (JSON and YAML). This table is the distillation result of averaging all the estimated true performance scores across models and
benchmarks in \Cref{tab:mapping-results}.
}}
\label{tab:mapping-values}
\end{table*}

We select Gemma, Mistral, and ChatGPT for our evaluation, using two prompting techniques: Zero-shot (ZS) and Zero-shot Chain-of-Thought (ZS-CoT) (\Cref{sec:mapping}). These models are tested on the SciREX dataset across three difficulty levels: Easy, Medium, and Hard. Detailed prompt specifications are provided in \Cref{appdx:mapping-prompt-details}. We calculate $FI_C, SysE, EstTrueE$ with the results presented in \Cref{tab:mapping-results}. Furthermore, \Cref{tab:mapping-values}, \Cref{tab:mapping-fi-among-formats} and \Cref{tab:mapping-cot-vs-non-cot} are distilled from \Cref{tab:mapping-results}:

\begin{enumerate}
    \item \Cref{tab:mapping-values}. For each model, we average its $EstTrueE$ performance overall benchmarks and prompting techniques. For each task, we average the $EstTrueE$ scores overall models and prompting techniques. This table is numerical results of \Cref{fig:mapping_main} and discussed in \Cref{ssec:mapping-findings}.
    \item \Cref{tab:mapping-fi-among-formats}. The purpose of this table is to compare the FI scores across formats. We average all the FI scores across models and tasks. 
    \item \Cref{tab:mapping-cot-vs-non-cot}. The purpose of this table is to see whether CoT \cite{wei2022chain} mitigates format bias. We average all the $EstTrueE$ scores over all models and benchmarks for each ZS and ZS-CoT prompting method. 
\end{enumerate}


\paragraph{Format instruction-following bias.}
From this \cref{tab:mapping-fi-among-formats}, noticeably, there is a significant difference in the FI scores of two mapping formats (``JSON'' and ``YAML''). The average FI scores of the ``JSON'' mapping format is $60.98\%$, which is much higher than ``YAML'' mapping format ($43.96\%$). Overall, among models, ChatGPT yields the best result of $90.40\%$. Mistral fails to follow ``YAML'' format instruction, e.g. for the question in dataset \textbf{SciREX Easy} \Cref{appdx:dataset} with detailed promt as in \Cref{appdx:mapping-prompt-details}

\begin{tcolorbox}[colback=white]
    \texttt{Extract the entities reflecting the tasks in the following document:}\\
    \\
    Document: \texttt{Published as a conference paper at ICLR 2017 QUERY - REDUCTION NETWORKS FOR QUESTION ANSWERING.\\
    \\
    \textcolor{blue}{Your output must be in YAML format: `````` Task: [...] ''''''. Wrap your final answer by <ANSWER> and </ANSWER>.}} \\
    \\
    \texttt{\textbf{Mistral:} \textcolor{red}{```yaml\\
\texttt{- TASK\_1: Prepare conference paper}\\
\texttt{- TASK\_2: Submit conference paper to ICLR 2017}\\
\texttt{- TASK\_3: Research and develop Query-Reduction Networks for Question Answering}\\
'''\\
\texttt{<ANSWER>}\\
```\texttt{yaml}\\
\texttt{- TASK\_1: Prepare conference paper}\\
\texttt{- TASK\_2: Submit conference paper to ICLR 2017}\\
\texttt{- TASK\_3: Research and develop Query-Reduction Networks for Question Answering}'''\\
\texttt{</ANSWER>}}}.
\end{tcolorbox}

\begin{table*}[!htp]
\centering
\footnotesize
\scalebox{1}{
\begin{tabular}{c|cc|cc}
\toprule
\textbf{Model} & \textbf{JSON} & \textbf{YAML} & \textbf{Avg.} & \textbf{$BiasF_o$}\\
\midrule
Gemma & 20.71 & 12.62 & 16.66 & \textbf{16.34} \\
Mistral & 82.82 & 5.13 & 43.98 & 1509.00\\
Llama & 45.70 & 71.97 & 58.84 & 172.46\\
ChatGPT & \textbf{94.69} & \textbf{86.11} & \textbf{90.40} & 18.41 \\
\midrule
\textbf{Avg.} & 60.98 & 43.96 & 52.47 & 72.45\\
\bottomrule
\end{tabular}}
\caption{\small{Avg FI scores over benchmarks and models across 2 widely used mapping formats (JSON and YAML). This table is the distillation result of averaging all the FI scores across models and benchmarks in \Cref{tab:mapping-results}.}}
\label{tab:mapping-fi-among-formats}
\end{table*}

\paragraph{ZS versus ZS-CoT.}

\begin{table*}[!htp]
\centering
\footnotesize
\scalebox{1}{
\begin{tabular}{lcc|c}
\toprule
& \textbf{JSON} & \textbf{YAML} & \textbf{$BiasF_o$}\\
\midrule
Zero-shot & 28.02 & 16.79 & \textbf{31.55} \\
Zero-shot Chain-of-Thought & 26.01 & 15.29 & 28.73 \\
\bottomrule
\end{tabular}}
\caption{\small{Avg ZS and ZS-CoT scores over benchmarks and models across 2 widely used mapping formats (JSON and YAML). This table is the distillation results across models and benchmarks in \Cref{tab:mapping-results}.}}
\label{tab:mapping-cot-vs-non-cot}
\end{table*}

From \cref{tab:mapping-cot-vs-non-cot}, it is evident that the performance of ZS prompting surpasses that of ZS-CoT for both formats. Upon comparing the $BiasF_o$ across prompting techniques, we conclude that CoT \cite{wei2022chain} does not mitigate format bias.


\paragraph{Reliability of the results.}
From \cref{tab:mapping-results}, $77\%$ of the $EstTrue$ results are reliable. The reliability of the results in the mapping output format shows variability across different models and formats. Noticeably, ``YAML'' mapping format results are less reliable than ``JSON'' ones. On the other hand, ChatGPT illustrates its high reliability in all mapping formats while Mistral-7B-it and Gemma-7B-it are opposite, and all the results in the ``YAML'' mapping format of these models are unreliable. 

\subsection{Mitigating Format Bias Results}

\begin{table*}[!htp]
\centering
\footnotesize
\scalebox{.65}{
\begin{tabular}{l|lccccccc|cc}
\toprule 
\textbf{Index} & \textbf{Wrapping type} & \textbf{Special character} & \textbf{Bolding} &  \textbf{Italicizing} & \textbf{Brackets} & \textbf{Parentheses} & \textbf{Placeholder} & \textbf{Quoting} & \textbf{Avg.} & \textbf{\textbf{$BiasF_o$} (Var)} \\
\midrule
& & & \textbf{No demo (Zero-shot)} & & & \\
1& ChatGPT (EstTrue-Acc) & 54.63 & 67.39 & 44.76 & 59.79 & 57.82 & 66.23 & 68.28 &  & 235.33  \\
2&ChatGPT (Systematic) & 48.54  & 66.59 & 38.24 & 31.65 & 28.54 & 63.88 & 26.72 &  & 532.75\\
3&ChatGPT (FI) & 88.84  & 98.80 & 85.43 & 52.93 & 49.36 & 96.44 & 39.13  & 72.99 & 61.12 \\
\midrule
& & & \textbf{Repeat format prompt thrice} & & & \\
4&ChatGPT (EstTrue-Acc) & 60.09 & 67.88 & 55.65 & 61.99 & 63.71 & 30.31 & 68.28 &  & 146.79 \\
5&ChatGPT (Systematic) & 56.65  & 66.98 & 49.93 & 35.74 & 51.63 & 2.85 & 33.13 &  & 377.66 \\
6&ChatGPT (FI) & 94.26  & 98.67 & 89.71 & 57.65 & 81.03 & 9.40 & 48.52  & 68.46 & 884.34 \\
\midrule
& & & \textbf{1 demo} & & & \\
7&ChatGPT (EstTrue-Acc) & 55.12 & 65.08 & 47.18 & 52.23 & 56.13 & 65.92 & 63.60 &  & 172.69 \\
8&ChatGPT (Systematic) & 50.54 & 64.49 & 43.98 & 40.02 & 31.02 & 62.19 & 28.10 &  & 397.62 \\
9&ChatGPT (FI) & 91.68 & 99.09 & 93.22 & 76.61 & 55.26 & 94.34 & 44.18 & 79.20 &  43.75 \\
\midrule
& & & \textbf{5 demos} & & & \\
10&ChatGPT (EstTrue-Acc) & 51.77 & 58.30 & 45.21 & 46.79 & 52.52 & 62.84 & 55.24 &  &  111.78 \\
11&ChatGPT (Systematic) & 51.18 & 56.66 &  40.69 & 41.36 & 39.78 & 60.88 & 27.72 &  & 259.37 \\
12&ChatGPT (FI) & 98.85 & 97.19 & 90.01 & 88.39 & 75.74 & 96.88 & 50.18 & 85.32 & 32.93 \\
\midrule
& & & \textbf{Finetuned} & & & \\
13&ChatGPT (EstTrue-Acc) & 74.02 & 74.73 & 71.53 & 73.88 & 74.09 & 74.27 & 74.19 &  & \textbf{0.71} \\
14&ChatGPT (Systematic) & 73.99 & 74.11 & 71.52 & 73.66 & 73.47 & 74.15 & 73.70 &  & \textbf{0.11} \\
15&ChatGPT (FI) & 99.96 & 99.17 & 99.98 & 99.69 & 99.16 & 99.83 & 99.33 & \textbf{99.59} & \textbf{0.93} \\
\bottomrule
\end{tabular}}
\scalebox{.65}{
\begin{tabular}{l|lccccccc|cc}
\toprule 
\textbf{Index} & \textbf{Wrapping type} & \textbf{Special character} & \textbf{Bolding} &  \textbf{Italicizing} & \textbf{Brackets} & \textbf{Parentheses} & \textbf{Placeholder} & \textbf{Quoting} & \textbf{Avg.} & \textbf{\textbf{$BiasF_o$} (Var)} \\
\midrule
& & & \textbf{No demo (Zero-shot)} & & & \\
16& Gemma-2B  (EstTrue-Acc) & 30.64 & 6.88 & 1.90 & 0.0 & 0.0  & 8.41 & 0.0  & & 104.81  \\
17&Gemma-2B  (Systematic)  & 23.26 & 4.32 & 0.78 & 0.0 & 0.0 & 8.32 & 0.0  & & 63.00 \\
18&Gemma-2B  (FI) & 76.24 & 62.83 & 41.06 & 0.0 & 0.0 & 98.91 & 0.0  & 39.86 & 1443.70\\
\midrule
& & & \textbf{Repeat format prompt quintice} & & & \\
19&Gemma-2B  (EstTrue-Acc) & 58.40  & 0.61  & 32.26 & 0.00 & 0.00 & 55.52 & 0.00 &  & 636.50 \\
20&Gemma-2B  (Systematic) & 56.00 & 0.40 & 14.00 & 0.00 & 0.00 & 54.00 & 0.00 &  & 576.77 \\
21&Gemma-2B  (FI) & 95.89 & 65.35 & 43.40 & 0.00 & 0.00 & 97.27 & 0.00 & 43.13 & 1684.16 \\
\midrule
& & & \textbf{1 demo} & & & \\
22&Gemma-2B  (EstTrue-Acc) & 35.57 & 34.45 & 34.84 & 31.76 & 0.00 & 26.35 & 34.34 &  & 140.95 \\
23&Gemma-2B  (Systematic) & 32.03 & 34.45 & 25.25 & 26.81 & 0.00 & 6.07 & 27.49 &  & 84.32 \\
24&Gemma-2B  (FI) & 90.03 & 99.98 & 72.47 & 84.41 & 0.00 & 23.03 & 80.05 & 74.99 & 611.87 \\
\midrule
& & & \textbf{5 demos} & & & \\
25&Gemma-2B  (EstTrue-Acc) & 38.56 & 38.90 & 38.49 & 36.72 & 36.52 & 38.97 & 37.25 &  & \textbf{0.94} \\
26&Gemma-2B  (Systematic) & 38.56 & 38.90 & 38.49 & 36.72 & 36.08 & 38.29 & 37.20 & & \textbf{0.99} \\
27&Gemma-2B  (FI) & 100.00 & 100.00 & 100.00 & 100.00 & 98.78 & 98.26 & 99.86 & \textbf{99.57} & \textbf{0.45} \\
\bottomrule
\end{tabular}
}
\caption{\small{Supplementing demonstrations, repeating format instructions, and extra fine-tuning with formats' data reduce format bias. Performance of ChatGPT and Gemma-2B-it (\textbf{without CoT}) on MMLU. All results are reliably measured by \Cref{theorem:theorem1}.}}
\label{tab:more-demonstrations}
\end{table*}

In this section, we present the numerical results of our proposed techniques for mitigating format biases using ChatGPT on MMLU, as shown in \Cref{tab:more-demonstrations} indexes 1-15.

\begin{enumerate}
\item \textbf{Demonstrations with formats reduce bias (Indexes 7-12).} From \Cref{tab:more-demonstrations} indexes 7-12, we observe that using demonstrations with formats generally increases the average of FI scores, from $72.99\%$ without any demonstration (index 3), to $79.20\%$ with using one demonstration and $85.32\%$ with using $5$ demonstrations. Moreover, we find that the performance does not scale linearly with the FI score, indicating that simply increasing the FI score does not necessarily improve the models' performance or reduce format biases.
\item \textbf{Repeating format instructions reduces format bias (Indexes 4-6).} From \Cref{tab:more-demonstrations} index 6, most of the formats, repeating the format instruction can increase the FI score (compared to index 3), except for the ``Placeholder''. Manual investigation reveals that repeatedly using the ``Placeholder'' format confuses the model about the actual location of the placeholder, leading to the model omitting the format. Nevertheless, this strategy generally reduces the format bias by decreasing the variance of results from formats other than ``Placeholder'', leading to overall reduction.
\item \textbf{Fine-tuning with additional format data can eliminate format bias (Indexes 13-15).} Finetuning mostly eliminates the format bias problem of the LLM with the bias score only $0.71\%^2$ from \Cref{tab:more-demonstrations} indexes 13-15, while increasing the average FI score up to almost perfect with $99.59\%$. This demonstrates that finetuning can help LLMs become more familiar with format tokens and requirements, reducing bias towards different formats.   
\end{enumerate}

For \textbf{Gemma-2B-it} model indexed 16-27, we observe that: 
\begin{enumerate}
    \item \textbf{Repeating format instruction (Indexes 19-21).} This approach does not resolve Gemma-2B-it's inability to adhere to the ``Brackets'' and ``Parentheses'' instructions. However, disregarding these two formats, the strategy effectively reduces format bias in the model's performance.
    \item \textbf{Demonstrations (Indexes 22-27).} Demonstrations significantly mitigate format bias, as evidenced by higher FI scores and reduced variance across format performances.
\end{enumerate}


\section{Experimental Details} \label{appdx:experimental-details}

\subsection{Dataset Details} \label{appdx:dataset}
We provide descriptions of all datasets we use in this paper.

\paragraph{MMLU \cite{hendrycks2020measuring}.} MMLU  is a benchmark for evaluating the performance of language models on Multiple Choices Question on a wide range of subjects across STEM, the humanities, social sciences, and other areas, testing the model's ability to understand and reason in diverse domains.

\paragraph{BBH \cite{suzgun-etal-2023-challenging}.} BBH is a MCQ dataset which includes a variety of challenging benchmarks that require advanced reasoning, comprehension, and other complex cognitive skills.

\paragraph{GSM8K \cite{cobbe2021gsm8k}.} GSM8K is a dataset of 8,000 math word problems designed for grade school students. The problems require not just basic arithmetic but also multi-step reasoning to solve.

\paragraph{HotpotQA \cite{yang-etal-2018-hotpotqa}.} HotpotQA is a question-answering dataset with a focus on multi-hop reasoning. It contains questions that require finding and combining information from multiple Wikipedia articles to derive the answer.

\paragraph{FairytaleQA \cite{xu-etal-2022-fantastic}.} FairytaleQA is a dataset designed for evaluating narrative comprehension, particularly in the context of children's fairytales. It includes questions that test understanding of characters, plots, and settings in fairytales.

\paragraph{SciDocsRR \cite{cohan-etal-2020-specter}.} SciDocsRR is a dataset for evaluating information retrieval systems, particularly in the scientific domain. It includes tasks like citation prediction, document classification, and other retrieval-based evaluations.

\paragraph{SemEval2017 \cite{augenstein-etal-2017-semeval}.} SemEval2017 is part of an ongoing series of evaluations for semantic analysis in natural language processing. It includes a wide range of tasks such as sentiment analysis, semantic textual similarity, and information extraction.

\paragraph{SciREX \cite{jain-etal-2020-scirex}.} SciREX is a dataset for evaluating models on the task of information extraction from scientific literature. It focuses on extracting entities, relations, and other structured information from research papers.


\subsection{Experimental Results} \label{appdx:experimental-results}

We present the hyperparameters setting for our experiments below.

\paragraph{Gemma-7B-it \cite{team2024gemma}.} For Gemma 7B-it, use the weights from Google and Huggingface\footnote{\url{https://huggingface.co/google/gemma-7b-it}}. We use Nucleus Sampling \cite{Holtzman2020The} as our decoding strategy with a $p$ value of $0.95$, a temperature value of $0.1$, and a window size of $1024$.

\paragraph{Mistral-7B-it-v0.2 \cite{jiang2023mistral}.} For Mistral 7B-it, use the weights from MistralAI and Huggingface\footnote{\url{https://huggingface.co/mistralai/Mistral-7B-Instruct-v0.2}}. We use Nucleus Sampling \cite{Holtzman2020The} as our decoding strategy with a $p$ value of $0.9$, and a window size of $1024$.

\paragraph{ChatGPT (gpt3.5-turbo-0125) \cite{openai2022chatgpt}.} For ChatGPT, we use the system role: ``You are helpful assistant!''. We set the ``max\_tokens'' to be 1024, ``top\_p=1'', ``frequency\_penalty=0'', ``presence\_penalty=0'', and the model mode is ``gpt3.5-turbo-0125''.

\paragraph{Datasets for finetuning ChatGPT and finetuning setups.} We preprocess the ``auxiliary\_train''\footnote{\url{https://huggingface.co/datasets/cais/mmlu/viewer/auxiliary_train}} dataset of MMLU \cite{hendrycks2020measuring}, resulting in the training set of $6500$ samples as discussed in \Cref{sec:mitigating-format-bias}. We preprocess a small, distinct validation set with the same ratio as the training set among formats ''20-20-40-40-50-20-50'', resulting in a total of $240$ samples for validation.

We use the default finetuning setup of OpenAI for ChatGPT. \textbf{Our finetuning costs $63.86$ US\$}.

\section{Prompting} \label{appdx:prompting}
\subsection{MCQ Prompt Details} \label{appdx:mcq-prompt-details}
The input for the models is the combination of the following components:
\begin{align*}
    \textbf{Input} = \textbf{\{non-CoT/CoT $\times$ Char./Text.\} Instruction} + \textbf{Question} + (\textbf{CoT Wrapping})
\end{align*}
where \textbf{non-CoT/CoT Instruction} shows that model uses Zero-shot or Chain-of-Thought, given that
\begin{itemize}
    \item \textbf{non-CoT $\times$ Char. Instruction} = ``Answer the following multiple-choice question by outputting only the designated character identifier.''
    \item \textbf{non-CoT $\times$ Text. Instruction} = ``Answer the following multiple-choice question by outputting the textual value of your choice without the character identifier without any textual description.'' 
    \item \textbf{CoT $\times$ Char. Instruction} = ``Answer the following multiple-choice question step-by-step by outputting only the designated character identifier.''
    \item \textbf{CoT $\times$ Text. Instruction} = ``Answer the following multiple-choice question step-by-step by outputting the textual value of your choice without the character identifier.''
\end{itemize}
\textbf{Question} is the main content of the task and \textbf{CoT Wrapping} is wrapping instruction if using CoT. i.e. \textbf{CoT Wrapping} = ``Wrap your final answer by <ANSWER> and </ANSWER>.''

\subsection{Wrapping Prompt Details} \label{appdx:wrapping-prompt-details}
The input for the models is the combination of the following components:
\begin{align*}
    \textbf{Input} = \textbf{non-CoT/CoT Instruction} + \textbf{Question} + \textbf{Wrapping Format Instruction}
\end{align*}
where \textbf{non-CoT/CoT Instruction} shows that model uses Zero-shot or Chain-of-Thought, given that
\begin{itemize}
    \item If MCQ task (MMLU,BBH)
        \begin{enumerate}
            \item \textbf{non-CoT Instruction} = ``Answer the following multiple-choice question by outputting only the designated character identifier.''
            \item \textbf{CoT Instruction} = ``Answer the following multiple-choice question step-by-step by outputting only the designated character identifier.''
        \end{enumerate}
    \item If generation task (GSM8K, HotpotQA, FairytaleQA)
        \begin{enumerate}
            \item \textbf{non-CoT Instruction} = ``Answer the following question.''
            \item \textbf{CoT Instruction} = ``Answer the following question step by step.''
        \end{enumerate}

\end{itemize}
\textbf{Question} is the main content of the task, and \textbf{Wrapping Format Instruction} is the format we want the model to output, detailed as
\begin{itemize}
    \item \textbf{Special Character wrapping} = ``Wrap your final answer by <ANSWER> and </ANSWER>.''
    \item \textbf{Bolding wrapping} = ``Wrap your final answer in bold by enclosing it with double asterisks.''
    \item \textbf{Italicizing wrapping} = ``Wrap your final answer in italics by enclosing it with single asterisks.''
    \item \textbf{Brackets wrapping} = ``Wrap your final answer using double square brackets.''
    \item \textbf{Parentheses wrapping} = ``Wrap your final answer using double parentheses.''
    \item \textbf{Placeholder wrapping} = ``Wrap your final answer by filling in the placeholder below: `So the answer is: \{\{placeholder\}\}'''
    \item \textbf{Quoting wrapping} = ``Wrap your final answer using triple quotation marks.''
\end{itemize}
\subsection{List Prompt Details} \label{appdx:list-prompt-details}

For dataset \textbf{SciDocsRR}, the input for the models is the combination of the following components:
\begin{align*}
    \textbf{Input} = &  \textbf{ Information} + \textbf{Requirement} + \textbf{List Format Instruction} \\ & + \textbf{non-CoT / CoT Instruction} + (\textbf{CoT Wrapping})
\end{align*}
where 
\begin{itemize}
    \item \textbf{Information} = \text{``Given a query, and a list of documents: \\
        \quad Topic: \textbf{Topic. } \\
        List of documents: \textbf{Samples}''}
    \item \textbf{Requirement} = ``You are required to transform the list of documents into a binary list of $1$ or $0$ where $1$ indicates the document brings useful information to the topic, and $0$ indicates the document does not bring useful information to the topic.''
    \item \textbf{List Format Instruction} includes four categories:
        \begin{enumerate}
            \item \textbf{Python} = ``Generate your binary list as a Python list''
            \item \textbf{Bullet} = ``Generate your binary list using bullet points''
            \item \textbf{Special Character} = ``Generate your binary list using \texttt{<SEP>} to separate elements''
            \item \textbf{New Line} = ``Generate your binary list such that each element is in a new line''
        \end{enumerate}
    \item \textbf{non-CoT / CoT Instruction} includes:
        \begin{enumerate}
            \item \textbf{non-CoT Instruction} = ``without any explanation.''
            \item \textbf{CoT Instruction} = ``step by step''
        \end{enumerate}
    \item \textbf{CoT Wrapping} = ``Wrap your final list by \texttt{<ANSWER>} and \texttt{</ANSWER>}.''
\end{itemize}

For dataset \textbf{SemEval2017}, the input for the models is the combination of the following components:
\begin{align*}
    \textbf{Input} = &  \textbf{ Requirement} + \textbf{Document} + \textbf{List Format Instruction} \\ & + \textbf{non-CoT / CoT Instruction} + (\textbf{CoT Wrapping})
\end{align*}
where
\begin{itemize}
    \item \textbf{Requirement} = ``Extract a list of keyphrases from the following document:''
    \item \textbf{Document} is the main content of the task.
    \item \textbf{List Format Instruction} includes four categories:
        \begin{enumerate}
            \item \textbf{Python} = ``Generate your binary list as a Python list''
            \item \textbf{Bullet} = ``Generate your binary list using bullet points''
            \item \textbf{Special Character} = ``Generate your binary list using \texttt{<SEP>} to separate elements''
            \item \textbf{New Line} = ``Generate your binary list such that each element is in a new line''
        \end{enumerate}
    \item \textbf{non-CoT / CoT Instruction} includes:
        \begin{enumerate}
            \item \textbf{non-CoT Instruction} = ``without any explanation.''
            \item \textbf{CoT Instruction} = ``step by step''
        \end{enumerate}
    \item \textbf{CoT Wrapping} = ``Wrap your final list by \texttt{<ANSWER>} and \texttt{</ANSWER>}.''
\end{itemize}

\subsection{Mapping Prompt Details} \label{appdx:mapping-prompt-details}
For all three datasets, we use the following formula for the input of the models
\begin{align*}
    \textbf{Input} = &  \textbf{ Requirement} + \textbf{Document} + \textbf{Mapping Format Instruction} + (\textbf{CoT Wrapping})
\end{align*}
where 
\begin{itemize}
    \item \textbf{Requirement} = ``Extract the entities reflecting the tasks in the following document:'' if using non-CoT model and ``Extract the entities reflecting the tasks in the following document step-by-step:'' if using CoT model
    \item \textbf{Document} is the main content of the task.
    \item \textbf{CoT Wrapping} = ``Wrap your final list by \texttt{<ANSWER>} and \texttt{</ANSWER>}.''
    \item \textbf{Mapping Format Instruction} starts with defining a specific format for the model and then instructs the model to follow. In detail, we have
        \begin{enumerate}
            \item For \textbf{Easy} dataset, we define:
            \begin{lstlisting}
        JSON_FORMAT = {
                        ''Task'': [...]
                       }
            \end{lstlisting}
            \begin{lstlisting}
        YAML_FORMAT = '''''' Task: [...] ''''''
            \end{lstlisting}
            Then
            \begin{itemize}
                \item \textbf{JSON Mapping} = ``Your output must be a Python dictionary with the key `Task' and value as a list of task name entities: \verb|{str(JSON_FORMAT)}|''
                \item \textbf{YAML Mapping} = ``Your output must be in YAML format: \verb|{str(YAML_FORMAT)}|''
            \end{itemize}

        \item For \textbf{Medium} dataset, we define:
            \begin{lstlisting}
        JSON_FORMAT = {
                        ''Task'': [...],
                        ''Method'': [...]
                        }
            \end{lstlisting}
            \begin{lstlisting} 
        YAML_FORMAT = ''''''
                    Task: [...]
                    Method: [...]
                    ''''''
            \end{lstlisting}
            Then
            \begin{itemize}
                \item \textbf{JSON Mapping} = ``Your output must be a Python dictionary with the keys `Task' and `Method', and value is a list of task name entities and method name entities: \verb|{str(JSON_FORMAT)}|''
                \item \textbf{YAML Mapping} = ``Your output must be in YAML format: \verb|{str(YAML_FORMAT)}|''
            \end{itemize}

        \item For \textbf{Hard} dataset, we define:
            \begin{lstlisting}
        JSON_FORMAT = {
                        ''Task'': [...],
                        ''Method'': [...],
                        ''Material'': [...],
                        ''Metric'': [...]
                        }
            \end{lstlisting}
            \begin{lstlisting}
        YAML_FORMAT = ''''''
                    Task: [...]
                    Method: [...]
                    Material: [...]
                    Metric: [...]
                    ''''''
            \end{lstlisting}
            Then
            \begin{itemize}
                \item \textbf{JSON Mapping} = ``Your output must be a Python dictionary with the keys are `Task', `Method', `Material', `Metric', and value is a list of task name entities, method name entities, material name entities, metric name entities: \verb|{str(JSON_FORMAT)}|''
                \item \textbf{YAML Mapping} = ``Your output must be in YAML format: \verb|{str(YAML_FORMAT)}|''
            \end{itemize}
    \end{enumerate}
\end{itemize}





\end{document}